\documentclass[preprint,12pt]{elsarticle}

\usepackage{epsfig}

\usepackage[margin=2.5cm]{geometry}

\usepackage{amssymb}
\usepackage{amsmath}
\usepackage{amsthm}
\usepackage{xcolor}
\usepackage{booktabs}

\usepackage{graphicx}
\usepackage{setspace}
\usepackage{float}
\usepackage{lmodern}

\newtheorem{definition}{Definition}
\newtheorem*{remark}{Remark}

\newtheorem{problem}{Problem}
\usepackage{makecell}
\usepackage{multirow}

\journal{Nuclear Physics B}

\begin{document}

\begin{frontmatter}



\title{Hierarchical Forecast Reconciliation for Urban Rail Transit Demand Prediction under Operational Disruptions} 


\author[1]{Dang Viet Anh Nguyen} 
\author[1]{Alma Fazlagic} 
\author[1]{Kristine Pryds Loft}
\author[1]{Filipe Rodrigues}

\affiliation[1]{organization={Department of Technology, Management, and Economics, Technical University of Denmark (DTU)},
            postcode={2800 Kongens Lyngby}, 
            country={Denmark}}


\begin{abstract}
Accurate and coherent passenger demand forecasting is essential for the
operational management of Urban Rail Transit (URT) systems. Passenger demand
has a hierarchical structure in which origin--destination (OD) flows aggregate
to station-level inflows and outflows through conservation constraints.
However, independently generated station- and OD-level forecasts can violate
these constraints, resulting in cross-level inconsistency and limiting the
effective integration of information across the hierarchy. This paper develops a hierarchical forecast reconciliation framework for joint station-level and OD-level demand prediction. A neural Fully Connected Reconciler (FCR) maps incoherent base forecasts to coherent predictions, with exact structural consistency guaranteed by construction. We benchmark FCR against classical and machine-learning-based reconciliation methods using one year of Rejsekort smart-card data and Banedanmark operational data from a 12-station subnetwork of the Copenhagen S-train system, considering one-step, multi-step, and disruption forecasting settings. We additionally compare against a joint multi-task station--OD baseline trained under a matched parameter budget, in which a soft aggregation-consistency penalty reduces but does not eliminate cross-level disagreement. Reconciliation improves the principal aggregate OD-accuracy metrics while enforcing exact coherence. Under standard conditions, FCR performs competitively with the evaluated statistical methods, while a counterfactual oracle experiment using observed target-interval station
demand as the station-level input reduces OD MSE by approximately 34\%. Under train-delay and cancellation conditions, reconciliation continues to
improve OD accuracy, with FCR yielding larger point reductions than
MinT-Sample in the disruption-affected subsets. Disrupted intervals exhibit greater demand variability and cross-level forecast disagreement, and larger disagreement is positively associated with greater reconciliation gains. Within the studied subnetwork, these results demonstrate that hierarchical reconciliation provides structural coherence and useful cross-level correction under both regular and disrupted operating conditions.
\end{abstract}




\end{frontmatter}



\section{Introduction}
\label{sec1}

Urban Rail Transit (URT) systems form the backbone of mobility in modern cities, serving millions of daily journeys across increasingly interconnected networks. As urban populations grow, accurate short-term passenger demand forecasting has become essential for effective operations, underpinning critical decisions in timetable planning, rolling-stock allocation, crowd management, and disruption response. Yet rising travel demand continues to intensify recurrent operational challenges, including platform overcrowding, unstable passenger flows, and capacity bottlenecks, all of which threaten the reliability and resilience of urban transit systems \cite{Li2024Long-TermInformer, Lv2024AnMechanism}. These pressures have motivated substantial interest in data-driven and deep learning approaches to passenger demand forecasting \cite{Halyal2022ForecastingData}.

A defining characteristic of URT demand is its hierarchical structure. Passenger flows can be represented at multiple levels of granularity, from aggregated station-level inflows and outflows to fine-grained origin--destination (OD) flows between station pairs. These levels are linked by inherent aggregation constraints: the sum of outbound OD flows from a station must equal its total outflow, with an analogous relationship holding for inflows. In practice, station-level and OD-level forecasts are often generated by separate models optimised for different prediction tasks, without exploiting these structural relationships. Yet the two forecasting levels contain complementary information: station-level models characterise aggregate temporal demand patterns, whereas OD-level models preserve detailed spatial flow distributions. The known aggregation structure therefore provides a principled mechanism for combining information across levels to improve forecasting accuracy. Forecast reconciliation performs this cross-level information integration while ensuring that the resulting station- and OD-level predictions satisfy the network conservation constraints \cite{Spiliotis2021HierarchicalML, Wickramasuriya2019OptimalReconciliation}. This capability is particularly relevant under operational disruptions, when shifted and more variable demand conditions may affect the two base forecasting models differently and increase the potential benefit of cross-level reconciliation.

Reconciliation methods of this kind have been extensively studied in hierarchical time-series forecasting, particularly in economics, retail, and energy \cite{Hyndman2011OptimalSeries, Athanasopoulos2024ForecastReview}. Yet their application to urban mobility forecasting, and specifically to reconciling OD flows with station-level demand, remains largely unexplored. URT systems introduce a distinct set of challenges that make this problem non-trivial: directional OD structures, quadratically growing and sparse OD matrices, partial observability of trip destinations in smart-card data, and strong non-stationarity under operational disruptions. These characteristics make coherent forecasting both technically demanding and operationally significant.

This study investigates whether hierarchical forecast reconciliation can improve forecasting accuracy while ensuring exact coherence in deep learning-based URT demand prediction. We propose a two-level framework combining a sequence-to-sequence model, which jointly predicts station outflow and inflow, with a graph neural network for OD-level forecasting, with exact hierarchical coherence enforced at inference time through a fixed summing matrix $\mathbf{S}$ that encodes the conservation constraints of the network. Building on these base models, we evaluate neural and statistical reconciliation methods using real-world Rejsekort smart-card data from the Copenhagen S-train network. The one-step evaluation includes FCR, OLS, WLS, and several MinT variants, while the multi-step evaluation compares the shared-weight extension of FCR, denoted S-FCR, with horizon-specific MinT-Sample reconciliation.

A central motivation of this work is to determine whether hierarchical reconciliation remains effective when operational disruptions alter passenger demand patterns and affect station- and OD-level forecasts differently. We evaluate reconciliation under train cancellations and delays exceeding 60 seconds, and analyse both the associated differences in demand characteristics and the coherence violations between the independently generated forecasts at the two levels. The results show that reconciliation continues to improve OD forecasting accuracy under disrupted conditions. Moreover, representative disrupted conditions exhibit larger base coherence violations than undisrupted conditions, and these violations are positively, although moderately, associated with greater reductions in OD forecasting error. These findings suggest that reconciliation is particularly useful when the station- and OD-level base forecasts provide divergent but complementary information.

The main contributions of this work are as follows:

\begin{itemize}

\item We develop a hierarchical deep learning forecasting framework that jointly integrates station-level and OD-level demand prediction, with exact coherence guaranteed by construction regardless of the base forecasting architectures used.

\item We provide a systematic comparison of neural and classical statistical reconciliation methods. The learned reconciler performs competitively under standard conditions
and yields larger point reductions than MinT-Sample in the representative
disruption-affected subsets, while its shared-weight multi-step extension
(S-FCR) maintains lower OD, station, and full-hierarchy MSE than
horizon-specific MinT-Sample across all six horizons. We further compare against a joint multi-task station--OD baseline trained under a matched parameter budget, which reduces but does not eliminate cross-level disagreement, isolating the distinction between soft consistency mechanisms and exact reconciliation.

\item We conduct an empirical evaluation using real-world Rejsekort smart-card and
Banedanmark operational data from the Copenhagen S-train network, showing
that reconciliation gains are often larger under representative disruption
conditions and are positively associated with greater coherence violations
between station- and OD-level base forecasts.

\end{itemize}

\section{Related Work}
\label{sec:related}

Accurate and coherent forecasting across multiple levels of aggregation is essential in many operational domains, motivating extensive research on hierarchical and grouped time-series reconciliation. Classical reconciliation methods adjust independently generated base forecasts to satisfy known aggregation constraints, thereby improving interpretability and downstream decision-making. Such approaches are now well established in economics, retail, energy, and tourism forecasting \cite{Hyndman2011OptimalSeries, Athanasopoulos2024ForecastReview}. Despite their maturity, the application of reconciliation to transportation systems, and particularly to high-dimensional, directional OD flows, remains comparatively underexplored. URT systems present unique challenges including severe data sparsity, partial observability of OD matrices, and sharp temporal irregularities during disruptions, characteristics that differ substantially from the structures for which classical reconciliation was designed.

\subsection{Hierarchical Forecast Reconciliation}

Forecast reconciliation addresses the problem of adjusting a set of independently generated forecasts so that they satisfy the aggregation constraints of a hierarchy. Given a summing matrix $\mathbf{S}$ that encodes how bottom-level series aggregate to higher levels, reconciliation produces a revised set of forecasts $\tilde{\mathbf{y}}_h = \mathbf{S}\mathbf{G}_h\hat{\mathbf{y}}_h$, where $\mathbf{G}_h$ maps the full set of base forecasts to the bottom level, and $\mathbf{S}$ re-aggregates to produce coherent forecasts at all levels \cite{Hyndman2011OptimalSeries}. The choice of $\mathbf{G}_h$ defines the reconciliation strategy, and a large family of methods can be expressed within this common framework \cite{Hollyman2021UnderstandingReconciliation}.

The earliest approaches generate forecasts at a single level and propagate them through the hierarchy. Bottom-up methods forecast each bottom-level series independently and aggregate upward, preserving fine-grained information but susceptible to noise at the most disaggregate levels. Top-down methods forecast only the top-level series and disaggregate using historical or forecast-based proportions; they are more robust to noise but discard lower-level information and can introduce bias in the reconciled forecasts \cite{Athanasopoulos2024ForecastReview}. Middle-out methods occupy an intermediate position, forecasting at some intermediate level and propagating both upward and downward, but share the general limitation of ignoring information from all other levels. In general, all single-level approaches are limited by their reliance on a single level of the hierarchy, ignoring potentially valuable information contained at others.

Optimal combination methods overcome this limitation by using forecasts from all levels jointly. \cite{Hyndman2011OptimalSeries} showed that bottom-up, top-down, and middle-out methods are all special cases of the general linear reconciliation framework, and proposed an OLS solution that treats all series equally regardless of relative forecast accuracy. \cite{Hyndman2016WLS} addressed this by proposing a weighted least squares (WLS) solution that scales each series by the inverse variance of its base forecast errors. The minimum trace (MinT) reconciliation of \cite{Wickramasuriya2019OptimalReconciliation} extended this further by minimising the sum of variances of all reconciled forecast errors simultaneously, yielding a solution that subsumes OLS and WLS as special cases. In practice, the error covariance matrix required by MinT must be estimated from data, which is challenging when the number of series is large relative to available time periods, and shrinkage estimators are therefore commonly used \cite{Wickramasuriya2019OptimalReconciliation}. A complementary perspective is offered by \cite{Hollyman2021UnderstandingReconciliation}, who show that reconciliation is fundamentally a form of forecast combination: the aggregation constraints generate implicit alternative forecasts for each series, and reconciliation combines these with the direct base forecasts. This viewpoint explains why reconciliation gains tend to be largest precisely where forecasting is most difficult, at the noisiest and most disaggregate levels, where implicit forecasts derived from higher-level series provide the most useful correction.

Beyond classical linear methods, a growing body of work replaces the fixed projection of reconciliation with learned mappings capable of capturing non-linear cross-level dependencies. These approaches broadly fall into three categories. The first uses ML or DL models only to generate improved base forecasts, which are then reconciled using a classical method; \cite{Spiliotis2021HierarchicalML} demonstrate that replacing linear base forecasters with random forests and gradient boosting machines yields superior reconciled forecasts. The second category learns the reconciliation mapping itself; \cite{Burba2021Encoder} recast reconciliation as an encoder-decoder architecture, where a trainable encoder maps incoherent base forecasts to reconciled bottom-level predictions while a fixed summing matrix in the decoder guarantees exact structural consistency, an architecture that subsumes many existing strategies and demonstrates increasing gains for deeper hierarchies. The third category integrates forecasting and reconciliation end-to-end within a single training objective; \cite{Rangapuram2021EndToEnd} use a global neural network whose training loss incorporates the reconciled outputs directly, enabling joint optimisation for accuracy and coherence and capturing richer cross-series interactions than post-hoc reconciliation permits. A related line of work embeds hierarchical consistency as a soft regularisation penalty in the loss function \cite{Athanasopoulos2024ForecastReview}, though such approaches do not guarantee exact satisfaction of the aggregation constraints.

\subsection{Multi-Level Demand Prediction in Urban Rail Transit}

\label{sec:multilevel_urt}

A substantial body of research has developed deep learning models for short-term passenger-flow prediction at the station level, often using graph neural networks to capture spatial dependencies between stations
\cite{Wang2021HypergraphMetro,Liu2020PhysicalVirtual,Ma2019ParallelBiLSTM,
Bao2022AttentionMultiview,Fang2024DualView}.
These models have become increasingly sophisticated, incorporating multi-view graphs, adaptive adjacency matrices, recurrent structures, and attention mechanisms to represent complex spatio-temporal demand patterns. Nevertheless, station-level demand is generally treated as a single prediction target, without explicitly modelling its aggregation relationship with the finer-grained passenger flows from which it is formed. As noted by \cite{Lu2025AFlow}, conventional GNN-based passenger-flow models primarily capture spatial and temporal correlations between stations while overlooking the inter-level relationship between aggregate flows and their constituent sub-flows. In the terminology of the preceding subsection, these approaches can therefore be viewed as single-level forecasters: they operate at one level of the demand hierarchy without exploiting information available at other aggregation levels.

A second line of research introduces hierarchical demand structures explicitly. \cite{Lu2023MOHP-EC:Flow} propose the Multiobjective Hierarchical Prediction framework, which represents passenger demand at several aggregation levels, including ticket-type flows, station flows, and broader network-level groupings. Their results demonstrate that incorporating hierarchical relationships can improve both predictive accuracy and interpretability. Building on this idea, \cite{Lu2025AFlow} introduce IPF-HMGNN, which simultaneously predicts passenger flows by ticket type and aggregate station flows. A learnable projection mechanism coordinates the initial predictions so that the ticket-type components are consistent with the corresponding station-level total. The model achieves reductions in MAE and RMSE of up to 35\% and 36\%, respectively, on the Wuxi metro network. These studies establish the value of explicit hierarchical modelling in URT demand prediction. However, the hierarchies considered are primarily vertical structures within individual stations, relating categorical sub-flows to their station-level aggregates. They do not involve the sparse, directional, and network-wide OD matrix, whose dimensionality grows quadratically with the number of stations and whose entries may be only partially observable at prediction time.

A third and more directly related group jointly models multiple representations of station and OD demand through multi-task learning. HIAM \cite{liu2022online} jointly predicts inbound-based and outbound-based OD demand and introduces an OD-completion mechanism to address the partial observability of unfinished passenger trips, although station-level aggregates are not produced as separate prediction outputs. Multi-AFFN \cite{xu2023adaptive} uses adaptive feature fusion to transfer information across related station- and OD-demand prediction tasks. MTLMetro \cite{huang2024mtlmetro} jointly predicts station inflow, station outflow, and inbound- and outbound-based OD demand. Its knowledge-sharing mechanisms are derived from the relationships among these demand quantities: graph-based attention is used to propagate and distribute station-level information across the associated OD representations. AGSECIN \cite{zhao2025multi} similarly predicts station inflows, station outflows, and OD flows through interacting supply--demand representations. Concurrently with the present work, CATI \cite{yang2026cati} formulates station- and OD-level forecasting as structurally coupled tasks and introduces an aggregation-consistency regularisation term into the joint training objective.

These multi-task approaches demonstrate that cross-level information is
valuable and that passenger-flow conservation relationships provide useful
architectural inductive biases. Their primary objective, however, is joint
representation learning and prediction rather than output-level forecast
reconciliation. In MTLMetro, conservation relationships inform intermediate
message-passing operations, while the final forecasts are produced through
pooling, learnable feature fusion, and task-specific output layers. CATI
instead encourages cross-level agreement through a soft consistency penalty.
Such mechanisms may reduce disagreement between station-level and OD-level
predictions, but they do not impose an explicit equality constraint on the
final forecasts. Moreover, the remaining output-level inconsistency is not
reported as a dedicated evaluation quantity in the URT studies reviewed here,
particularly under operational disruptions.

\subsection{Positioning of This Work}

The preceding review highlights a gap between the forecast-reconciliation literature and existing multi-level URT demand prediction methods. Most URT studies predict station-level or OD-level demand independently
\cite{Wang2021HypergraphMetro,Liu2020PhysicalVirtual,Ma2019ParallelBiLSTM,
Bao2022AttentionMultiview,Fang2024DualView}, while recent multi-task models jointly predict both levels through shared representations, specialised message passing, or soft consistency penalties
\cite{xu2023adaptive,huang2024mtlmetro,zhao2025multi,yang2026cati}.
Other hierarchical approaches focus on ticket-type sub-flows and their station-level aggregates
\cite{Lu2023MOHP-EC:Flow,Lu2025AFlow}.
Among the studies reviewed here, none formulates station--OD forecasting as an optimal-combination reconciliation problem in which forecasts from both levels are jointly adjusted to satisfy the aggregation constraints exactly, nor is cross-level inconsistency evaluated as a dedicated quantity, including for the joint models that address the same structural relationship through softer mechanisms.

This gap is particularly relevant because URT OD demand is sparse,
directional, and potentially high-dimensional as the number of stations
increases, while disruptions further increase non-stationarity and
disagreement between prediction levels. The present work therefore combines station-level and OD-level base forecasts through statistical and neural reconciliation mappings and reconstructs the final forecasts using a fixed summing matrix, guaranteeing exact output-level coherence. The framework is evaluated under one-step, multi-step, and disruption forecasting settings to assess both predictive accuracy and cross-level consistency. To examine the distinction from joint multi-task prediction empirically rather than only conceptually, we additionally implement a joint station--OD network on the same data and parameter budget, in which the two levels are coupled during encoding and, in one variant, through a soft aggregation-consistency penalty (Sections~\ref{sec:joint_baseline} and~\ref{sec:joint_results}).

\section{Methodology}
\label{sec:method}

\subsection{Mathematical Framework and Problem Formulation}

Consider a URT network comprising $N$ stations, indexed by $i \in \mathcal{N} = \{1,\ldots,N\}$, observed over discrete time intervals $t \in \mathcal{T} = \{1,\ldots,T\}$. Passenger demand in such a network exhibits a natural hierarchical structure that we formalise through two complementary representations operating at distinct levels of granularity.
 

\begin{definition}[Station-level demand]
At the station level, let $x^{\mathrm{out}}_{i,t} \in \mathbb{Z}_+$
denote the total number of passengers departing from station
$i \in \mathcal{N}$ during interval $t \in \mathcal{T}$, and let
$x^{\mathrm{in}}_{i,t} \in \mathbb{Z}_+$ denote the destination-side
aggregate at station $i$ associated with passengers whose trips depart during
interval $t$. Thus, $x^{\mathrm{in}}_{i,t}$ is indexed by the departure
interval of the corresponding OD cohort rather than by the physical arrival
interval at station $i$. The outflow and inflow vectors across all stations are
\[
\mathbf{x}^{\mathrm{out}}_{t}
=
(x^{\mathrm{out}}_{1,t},\ldots,x^{\mathrm{out}}_{N,t})^\top
\in \mathbb{Z}^N_+,
\qquad
\mathbf{x}^{\mathrm{in}}_{t}
=
(x^{\mathrm{in}}_{1,t},\ldots,x^{\mathrm{in}}_{N,t})^\top
\in \mathbb{Z}^N_+.
\]
\end{definition}
 
\begin{definition}[OD-level demand]
At the OD level, let $y_{i,j,t} \in \mathbb{Z}_+$ denote the number of
passengers whose trips depart from origin station $i$ during interval $t$
and are destined for station $j \neq i$. Lexicographically ordering all
$N(N-1)$ origin--destination pairs $(i,j)$ with $i \neq j$ yields the
complete OD flow vector
$\mathbf{y}_t \in \mathbb{Z}^{N(N-1)}_+$, where the $k$-th entry
corresponds to the $k$-th pair in this ordering.
\end{definition}
 
\begin{remark}[Continuous relaxation]
Station outflows, inflows, and OD flows are all integer-valued passenger counts. Following standard practice in demand forecasting, we work throughout with the continuous relaxation $\mathbf{x}^{\mathrm{out}}_t, \mathbf{x}^{\mathrm{in}}_t \in \mathbb{R}^N_+$ and $\mathbf{y}_t \in \mathbb{R}^{N(N-1)}_+$, treating counts as real-valued aggregates over the discretisation interval. All subsequent analysis is conducted in this relaxed setting.
\end{remark}

These three representations are not independent: they are coupled through
the aggregation structure of the departure-indexed OD demand. For every
station $i \in \mathcal{N}$ and departure interval $t \in \mathcal{T}$,
the station outflow equals the sum of all OD flows originating from station
$i$, while the station inflow aggregate equals the sum of all OD flows
departing during interval $t$ whose destination is station $i$:
\begin{equation}
x^{\mathrm{out}}_{i,t}
=
\sum_{j \in \mathcal{N} \setminus \{i\}} y_{i,j,t},
\qquad
x^{\mathrm{in}}_{i,t}
=
\sum_{j \in \mathcal{N} \setminus \{i\}} y_{j,i,t}.
\label{eq:conservation}
\end{equation}
Both quantities are therefore exact row- and column-wise aggregates of the
same departure-indexed OD flow vector $\mathbf{y}_t$.
 
\begin{figure}
    \centering
    \includegraphics[width=\linewidth]{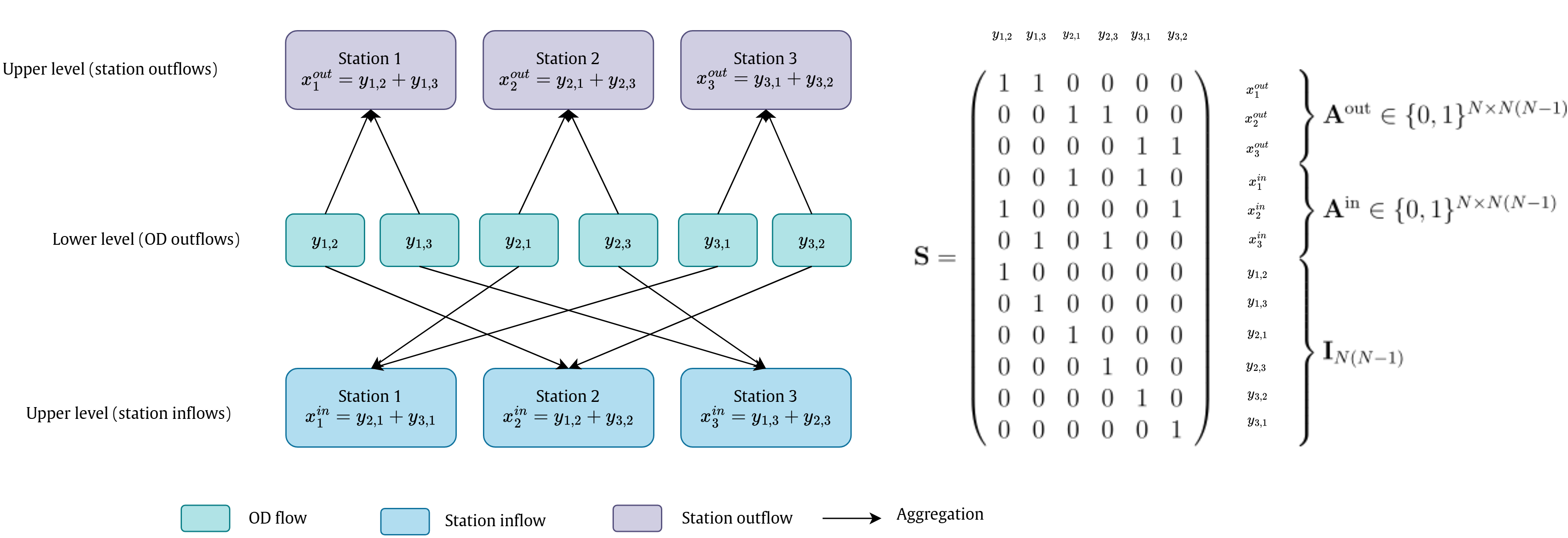}
    \caption{Two-level hierarchical demand structure for a URT network with
    $N=3$ stations. The station level comprises outflow nodes
    $x_i^{\mathrm{out}}$ and destination-side inflow nodes
    $x_i^{\mathrm{in}}$, both indexed by the departure interval of the
    corresponding OD cohort. The bottom level comprises $N(N-1)=6$ directed
    OD flow nodes $y_{i,j}$, one for each ordered station pair $(i,j)$ with
    $i\neq j$. Arrows indicate aggregation by origin to station outflows and
    by destination to station inflows. The summing matrix
    $\mathbf{S}\in\mathbb{R}^{12\times6}$ encodes this structure: the block
    $\mathbf{A}^{\mathrm{out}}\in\{0,1\}^{3\times6}$ aggregates OD flows by
    origin, the block
    $\mathbf{A}^{\mathrm{in}}\in\{0,1\}^{3\times6}$ aggregates them by
    destination, and $\mathbf{I}_6$ preserves the bottom-level OD flows.
    The complete demand vector $\mathbf{z}_t=\mathbf{S}\mathbf{y}_t$ stacks
    station outflows, station inflows, and OD flows, and is coherent if and
    only if $\mathbf{z}_t\in\mathrm{col}(\mathbf{S})$.}
    \label{fig:hierarchy}
\end{figure}

This pair of conservation constraints induces a hierarchical structure in which station-level outflows and inflows together form the upper level and disaggregated OD flows form the bottom level. To handle all three quantities jointly, we define the complete demand vector
$$\mathbf{z}_t = (\mathbf{x}^{\mathrm{out}}_{t}{}^\top, \mathbf{x}^{\mathrm{in}}_{t}{}^\top, \mathbf{y}_t^\top)^\top \in \mathbb{R}^M,$$
where $M = 2N + N(N-1)$ is the total number of series in the hierarchy. The relationship between the two levels can then be expressed compactly as:
\begin{equation}
\mathbf{z}_t = \mathbf{S}\mathbf{y}_t, \qquad \mathbf{S} = \begin{pmatrix} \mathbf{A}^{\mathrm{out}} \\ \mathbf{A}^{\mathrm{in}} \\ \mathbf{I}_{N(N-1)} \end{pmatrix} \in \mathbb{R}^{M \times N(N-1)},
\label{eq:summing}
\end{equation}
where the outflow aggregation matrix $\mathbf{A}^{\mathrm{out}} \in \{0,1\}^{N \times N(N-1)}$ has entry $A^{\mathrm{out}}_{i,k} = 1$ if OD pair $k$ originates from station $i$ and zero otherwise, the inflow aggregation matrix $\mathbf{A}^{\mathrm{in}} \in \{0,1\}^{N \times N(N-1)}$ has entry $A^{\mathrm{in}}_{i,k} = 1$ if OD pair $k$ terminates at station $i$ and zero otherwise, and $\mathbf{I}_{N(N-1)}$ is the identity matrix of order $N(N-1)$. The block $\mathbf{A}^{\mathrm{out}}$ encodes the first equality in equation~\eqref{eq:conservation}, the block $\mathbf{A}^{\mathrm{in}}$ encodes the second, and $\mathbf{I}_{N(N-1)}$ passes the OD flows through unchanged. We note that $\mathbf{z}_t$ takes non-negative values in practice; the non-negativity constraint on forecasts is treated as a practical desideratum handled separately from the structural coherence constraint. Figure~\ref{fig:hierarchy} illustrates this structure for a network of $N=3$ stations, together with the corresponding summing matrix $\mathbf{S}$.

In practice, forecasting models for station-level demand and OD flows are developed as separate systems, each optimised for its own prediction objective without access to the other's outputs. Even if trained jointly on shared data, the two models would in general still produce forecasts that violate the conservation constraints in equation~\eqref{eq:conservation}, since there is no mechanism within standard training objectives to enforce the structural relationships $\widehat{\mathbf{x}}^{\mathrm{out}}_{t+h} = \mathbf{A}^{\mathrm{out}}\widehat{\mathbf{y}}_{t+h}$ and $\widehat{\mathbf{x}}^{\mathrm{in}}_{t+h} = \mathbf{A}^{\mathrm{in}}\widehat{\mathbf{y}}_{t+h}$. We formalise this incompatibility as follows.

\begin{definition}[Base forecasts]
Let $\mathbf{X}_{t} \in \mathbb{R}^{2N \times \tau}$ denote the historical input matrix stacking station outflow and inflow series, and $\mathbf{Y}_{t} \in \mathbb{R}^{N(N-1) \times \tau}$ the historical OD flow series, where $\tau$ is the lookback window length. Let $\mathcal{F}_s: \mathbb{R}^{2N \times \tau} \to \mathbb{R}^{2N}$ and $\mathcal{F}_o: \mathbb{R}^{N(N-1) \times \tau} \to \mathbb{R}^{N(N-1)}$ denote the station-level and OD-level forecasting models. For forecast horizon $h \in \{1,\ldots,H\}$, these models jointly generate:
$$\bigl(\widehat{\mathbf{x}}^{\mathrm{out}}_{t+h},\, \widehat{\mathbf{x}}^{\mathrm{in}}_{t+h}\bigr) = \mathcal{F}_s(\mathbf{X}_{t}), \qquad \widehat{\mathbf{y}}_{t+h} = \mathcal{F}_o(\mathbf{Y}_{t}).$$
The concatenated base forecast vector is $\widehat{\mathbf{z}}_{t+h} = (\widehat{\mathbf{x}}^{\mathrm{out}}_{t+h}{}^\top, \widehat{\mathbf{x}}^{\mathrm{in}}_{t+h}{}^\top, \widehat{\mathbf{y}}_{t+h}{}^\top)^\top \in \mathbb{R}^M$.
\end{definition}

We denote the column space of $\mathbf{S}$ by $\mathrm{col}(\mathbf{S}) = \{\mathbf{S}\mathbf{v} : \mathbf{v} \in \mathbb{R}^{N(N-1)}\} \subseteq \mathbb{R}^M$, which we refer to as the \emph{coherent subspace} of the hierarchy. This is the set of all demand vectors that satisfy the conservation constraints of equation~\eqref{eq:conservation} exactly.
 
\begin{remark}[Partial observability of OD demand]
\label{rem:partial_obs}
In smart-card URT systems, $y_{i,j,t}$ denotes trips departing from origin
station $i$ during interval $t$ and destined for station $j$. The
corresponding station outflow is immediately available from tap-ins, whereas
the destination of a trip becomes known only when its tap-out is recorded.
Consequently, departure-indexed OD flows, and therefore their
destination-side aggregates
$x^{\mathrm{in}}_{j,t}=\sum_i y_{i,j,t}$, may remain incomplete at a
real-time prediction cutoff, particularly for journeys spanning several
aggregation intervals.
\end{remark}

This asymmetry in information availability means that the quantities used by
the station- and OD-level forecasting models have different observability
properties. Station outflow is available directly from tap-ins, whereas
departure-indexed OD demand and its destination-side inflow aggregate become
complete only as the corresponding tap-outs are observed. Independently
generated forecasts may therefore disagree even though the predicted
quantities are linked by exact aggregation constraints. Reconciliation
combines the information available at both levels and enforces these
constraints on the final predictions, but it does not directly recover
destinations that have not yet been observed.

\begin{definition}[Forecast coherence]
A forecast $\widehat{\mathbf{z}}_{t+h} \in \mathbb{R}^M$ is hierarchically coherent if and only if $\widehat{\mathbf{z}}_{t+h} \in \mathrm{col}(\mathbf{S})$, that is, if the predicted station outflows and inflows are exactly recoverable by aggregating the predicted OD flows:
$$\widehat{\mathbf{x}}^{\mathrm{out}}_{t+h} = \mathbf{A}^{\mathrm{out}}\widehat{\mathbf{y}}_{t+h}, \qquad \widehat{\mathbf{x}}^{\mathrm{in}}_{t+h} = \mathbf{A}^{\mathrm{in}}\widehat{\mathbf{y}}_{t+h}.$$
\end{definition}

Since $\mathcal{F}_s$ and $\mathcal{F}_o$ optimise separate objectives, the base forecasts will generally satisfy $\widehat{\mathbf{z}}_{t+h} \notin \mathrm{col}(\mathbf{S})$: the predicted station outflows and inflows will not equal the station-wise sums of the predicted OD flows. In this case, the direct station-level forecasts and the station aggregates
implied by the OD forecasts provide incompatible estimates of the same
passenger-flow quantities. Reconciliation combines the complementary
information contained in these forecasts while enforcing their structural
consistency. We refer to this situation as forecast incoherence. Reconciliation corrects this by projecting the incoherent base forecasts onto the coherent subspace while minimising the distortion introduced.

\begin{problem}[URT forecast reconciliation]
Given incoherent base forecasts $\widehat{\mathbf{z}}_{t+h} \notin \mathrm{col}(\mathbf{S})$ for a URT network, the forecast reconciliation problem is to find a reconciled forecast $\widetilde{\mathbf{z}}_{t+h}$ that simultaneously satisfies:
\begin{itemize}
    \item \textit{Coherence}: $\widetilde{\mathbf{z}}_{t+h} \in \mathrm{col}(\mathbf{S})$, equivalently $\widetilde{\mathbf{z}}_{t+h} = \mathbf{S}\widetilde{\mathbf{y}}_{t+h}$ for some $\widetilde{\mathbf{y}}_{t+h} \in \mathbb{R}^{N(N-1)}$, so that reconciled station outflows and destination-side inflow aggregates are consistent with the reconciled OD flows;
    \item \textit{Optimality}: $\widetilde{\mathbf{z}}_{t+h}$ minimises a loss functional $\mathcal{L}(\widetilde{\mathbf{z}}_{t+h}, \mathbf{z}_{t+h})$ subject to the coherence constraint, where $\mathcal{L}: \mathbb{R}^M \times \mathbb{R}^M \to \mathbb{R}_+$ is non-negative and satisfies $\mathcal{L}(\mathbf{z}, \mathbf{z}) = 0$.
\end{itemize}
\end{problem}

This problem admits a natural geometric interpretation: reconciliation seeks to project the incoherent forecast $\widehat{\mathbf{z}}_{t+h}$ onto $\mathrm{col}(\mathbf{S})$ according to the metric induced by $\mathcal{L}$. The choice of $\mathcal{L}$ determines which reconciliation method is obtained. We restrict attention to reconciliation mappings that admit the factorised form $\mathcal{R}(\widehat{\mathbf{z}}_{t+h}) = \mathbf{S}\mathbf{P}(\widehat{\mathbf{z}}_{t+h})$, where $\mathbf{P}: \mathbb{R}^M \to \mathbb{R}^{N(N-1)}$ maps base forecasts to reconciled bottom-level OD flows, and $\mathbf{S}$ re-aggregates to produce coherent station outflow and inflow forecasts. This factorisation covers all classical linear methods and the neural reconciler introduced in the following section.

When $\mathbf{P}$ is a fixed linear operator, $\mathbf{P}(\widehat{\mathbf{z}}) = \mathbf{G}\widehat{\mathbf{z}}$ for some $\mathbf{G} \in \mathbb{R}^{N(N-1) \times M}$, the reconciled forecast takes the form $\widetilde{\mathbf{z}}_{t+h} = \mathbf{S}\mathbf{G}\widehat{\mathbf{z}}_{t+h}$, recovering the standard linear reconciliation framework of \cite{Hyndman2011OptimalSeries}. Different choices of $\mathbf{G}$ yield the classical methods evaluated in this paper: bottom-up sets $\mathbf{G} = [\mathbf{0}_{N(N-1) \times 2N} \mid \mathbf{I}_{N(N-1)}]$, discarding the station-level forecasts entirely; WLS sets $\mathbf{G} = (\mathbf{S}^\top \boldsymbol{\Lambda}^{-1} \mathbf{S})^{-1} \mathbf{S}^\top \boldsymbol{\Lambda}^{-1}$ where $\boldsymbol{\Lambda} = \mathrm{diag}(\mathbf{W})$ is the diagonal of the forecast error covariance; and MinT sets $\mathbf{G} = (\mathbf{S}^\top \mathbf{W}^{-1} \mathbf{S})^{-1} \mathbf{S}^\top \mathbf{W}^{-1}$ where $\mathbf{W} = \mathrm{Var}(\mathbf{z}_{t+h} - \widehat{\mathbf{z}}_{t+h})$ is the full forecast error covariance matrix \citep{Wickramasuriya2019OptimalReconciliation}. However, linear reconciliation assumes that the cross-level relationships between station-level and OD-level demand are stationary and well-captured by second-order statistics, an assumption that is frequently violated in URT systems with high-dimensional and sparse OD matrices and non-stationary demand patterns under disruptions.

\subsection{Deep Learning Reconciliation for URT Station and OD Demand Prediction}

Classical linear reconciliation relies on a fixed projection matrix $\mathbf{G}$ estimated from historical forecast error covariances. In the URT station-OD hierarchy, this assumption faces two structural difficulties. First, the OD flow matrix is high-dimensional and sparse: with $N$ stations, the bottom level contains $N(N-1)$ OD series, and when this exceeds the number of training time steps, the error covariance matrix $\mathbf{W}$ is singular and must be approximated via shrinkage. Second, cross-level relationships between station-level and OD-level demand
may vary under disruptions and may therefore be difficult to capture using a
single fixed linear mapping estimated from historical forecast errors. These limitations motivate a learned reconciliation approach.
 
\begin{figure}
    \centering
    \includegraphics[width=\linewidth]{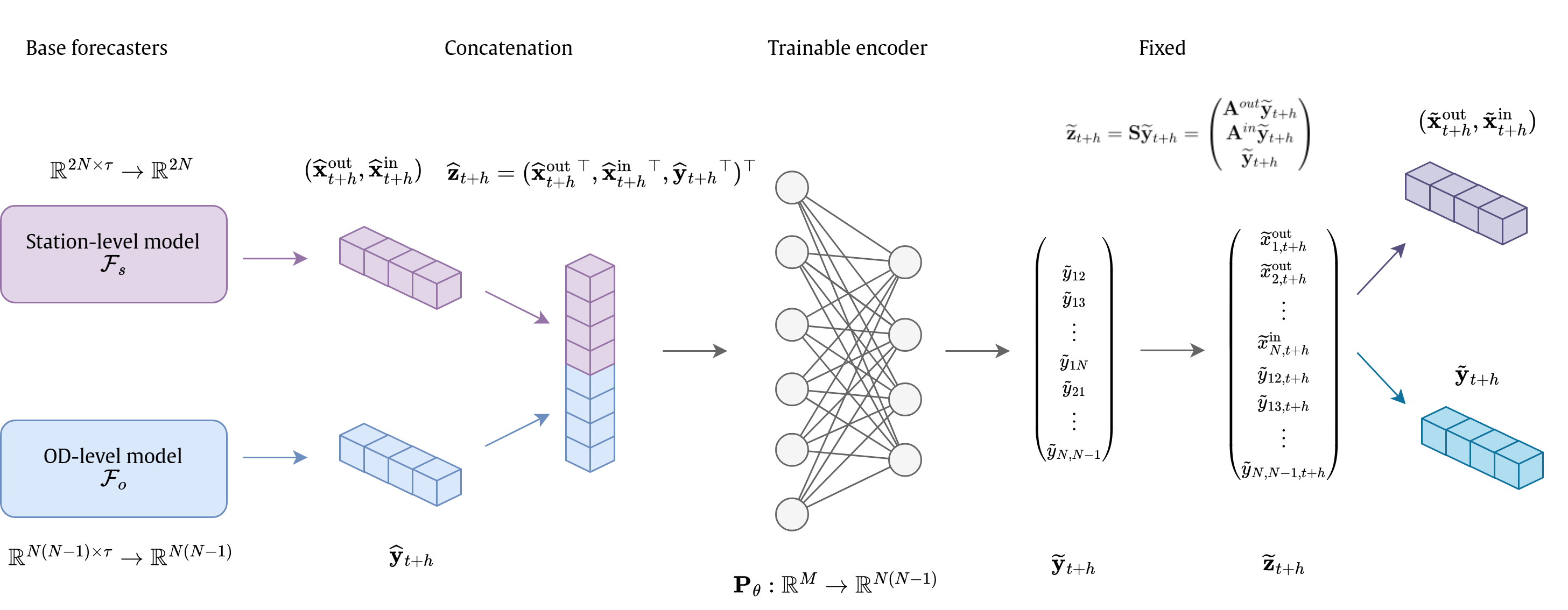}
    \caption{Encoder-decoder neural reconciliation architecture for URT station and OD demand forecasting, following \cite{Burba2021Encoder}. The station-level base forecaster jointly produces incoherent outflow and inflow predictions, and the OD-level base forecaster produces incoherent OD flow predictions; all three are concatenated into $\widehat{\mathbf{z}}_{t+h}$. The trainable encoder $\mathbf{P}_\theta$ maps this to reconciled bottom-level OD flows $\widetilde{\mathbf{y}}_{t+h}$. The fixed summing matrix $\mathbf{S}$ re-aggregates to recover coherent station outflow and inflow forecasts $\widetilde{\mathbf{x}}^{\mathrm{out}}_{t+h} = \mathbf{A}^{\mathrm{out}}\widetilde{\mathbf{y}}_{t+h}$ and $\widetilde{\mathbf{x}}^{\mathrm{in}}_{t+h} = \mathbf{A}^{\mathrm{in}}\widetilde{\mathbf{y}}_{t+h}$, guaranteeing $\widetilde{\mathbf{z}}_{t+h} \in \mathrm{col}(\mathbf{S})$ by construction. Classical linear methods (OLS, WLS, MinT) are special cases recovered when $\mathbf{P}_\theta$ is restricted to a linear layer.}
    \label{fig:reconciler}
\end{figure}

We adopt the encoder-decoder neural reconciliation framework of \cite{Burba2021Encoder}, in which the fixed linear projection $\mathbf{G}$ is replaced by a trainable neural network while the summing matrix $\mathbf{S}$ is retained as a fixed decoder, guaranteeing exact structural coherence by construction. We apply this framework as a post-hoc reconciliation step to the outputs of independently trained base forecasters. This design choice is appropriate because the station-level and OD-level models are architecturally distinct, consisting of a sequence-to-sequence network and a graph neural network, respectively, and are therefore best trained separately for their individual prediction objectives. This differs from the end-to-end paradigm of \cite{Rangapuram2021EndToEnd}, which requires a single global model to jointly represent all series in the hierarchy and is therefore less suited to settings with heterogeneous base forecasters. Figure~\ref{fig:reconciler} illustrates the full reconciliation pipeline, from independently trained base forecasters through the trainable encoder to the coherent output guaranteed by the fixed decoder $\mathbf{S}$.

\begin{definition}[Fully Connected Reconciler]
The neural reconciler is a feedforward network $\mathbf{P}_\theta: \mathbb{R}^M \to \mathbb{R}^{N(N-1)}$ with learnable parameters $\theta \in \Theta$, mapping the concatenated base forecasts of station outflows, station inflows, and OD flows to reconciled bottom-level OD predictions:
$$\widetilde{\mathbf{y}}_{t+h} = \mathbf{P}_\theta(\widehat{\mathbf{z}}_{t+h}).$$
The full hierarchically coherent forecast is then reconstructed via the structural summing matrix:
$$\widetilde{\mathbf{z}}_{t+h} = \mathbf{S}\widetilde{\mathbf{y}}_{t+h} = \begin{pmatrix} \mathbf{A}^{\mathrm{out}}\widetilde{\mathbf{y}}_{t+h} \\ \mathbf{A}^{\mathrm{in}}\widetilde{\mathbf{y}}_{t+h} \\ \widetilde{\mathbf{y}}_{t+h} \end{pmatrix},$$
so that $\widetilde{\mathbf{x}}^{\mathrm{out}}_{t+h} = \mathbf{A}^{\mathrm{out}}\widetilde{\mathbf{y}}_{t+h}$ and $\widetilde{\mathbf{x}}^{\mathrm{in}}_{t+h} = \mathbf{A}^{\mathrm{in}}\widetilde{\mathbf{y}}_{t+h}$ are the reconciled station outflow and inflow forecasts, and $\widetilde{\mathbf{y}}_{t+h}$ are the reconciled OD flow forecasts. By construction, $\widetilde{\mathbf{z}}_{t+h} \in \mathrm{col}(\mathbf{S})$ for any $\theta$, so both conservation constraints of equation~\eqref{eq:conservation} are satisfied exactly without any penalty term in the loss.
\end{definition}

\begin{remark}[Relationship to linear reconciliation]
When $\mathbf{P}_\theta$ is restricted to a linear function, $\mathbf{P}_\theta(\widehat{\mathbf{z}}) = \mathbf{G}\widehat{\mathbf{z}}$, the reconciler reduces to the classical methods: bottom-up, WLS, and MinT are all special cases of $\mathbf{G}$ \citep{Burba2021Encoder}. The neural reconciler therefore nests the entire family of linear methods and gains additional capacity to capture nonlinear cross-level dependencies when the linear assumption is violated.
\end{remark}

The parameters $\theta$ are estimated by minimising the empirical full-hierarchy mean squared error over the training set:
$$\widehat{\theta} = \arg\min_{\theta \in \Theta}\; \frac{1}{T_{\mathrm{train}}} \sum_{t=1}^{T_{\mathrm{train}}} \left\|\mathbf{z}_{t+h} - \mathbf{S}\mathbf{P}_\theta(\widehat{\mathbf{z}}_{t+h})\right\|_2^2,$$
which expands as:
$$= \frac{1}{T_{\mathrm{train}}} \sum_{t=1}^{T_{\mathrm{train}}} \left(\left\|\mathbf{x}^{\mathrm{out}}_{t+h} - \mathbf{A}^{\mathrm{out}}\widetilde{\mathbf{y}}_{t+h}\right\|_2^2 + \left\|\mathbf{x}^{\mathrm{in}}_{t+h} - \mathbf{A}^{\mathrm{in}}\widetilde{\mathbf{y}}_{t+h}\right\|_2^2 + \left\|\mathbf{y}_{t+h} - \widetilde{\mathbf{y}}_{t+h}\right\|_2^2\right),$$
jointly penalising deviations at the station outflow, station inflow, and OD flow levels. Following \cite{Burba2021Encoder}, we initialise $\mathbf{P}_\theta$ to approximate the bottom-up mapping, ensuring the untrained reconciler recovers the raw OD base forecasts and providing a stable starting point for optimisation.
 
For the experiments in this paper, $\mathbf{P}_{\theta}$ is implemented as a fully connected network whose parameter count grows as $O(N^4)$ because the number of OD flows grows quadratically with the number of stations. This architecture is tractable for the 12-station hierarchy considered here, but the present experiments do not establish its computational scalability to substantially larger networks. The contribution of the current study is therefore the formulation and evaluation of hierarchical reconciliation under operational disruptions, rather than the development of a large-scale reconciliation architecture. For larger networks, a structurally constrained variant in which each reconciled OD flow $\widetilde{y}_{i,j,t+h}$ depends only on its own base forecast $\widehat{y}_{i,j,t+h}$ together with the base forecasts for the corresponding origin outflow $\widehat{x}^{\mathrm{out}}_{i,t+h}$ and destination inflow $\widehat{x}^{\mathrm{in}}_{j,t+h}$ could reduce parameter growth to $O(N^2)$; we leave this extension to future work.
 
\begin{remark}[Universal approximation]
A multilayer perceptron with a non-polynomial activation function and
sufficient hidden units can approximate any continuous mapping on a compact
domain to arbitrary accuracy \citep{hornik1991approximation}. Accordingly,
the reconciler $\mathbf{P}_\theta$ has the representational capacity to model
nonlinear dependencies between station-level and OD-level forecasts that may
be difficult to capture using a single fixed linear reconciliation mapping.
This approximation property concerns representational capacity and does not,
by itself, guarantee improved forecasting performance or adaptation to
operational disruptions. The empirical evaluation in
Section~\ref{sec:experiments} assesses the performance of the learned
reconciliation mapping on the Copenhagen S-train case study.
\end{remark}

\section{Experiments}
\label{sec:experiments}
 
This section presents a systematic empirical evaluation of the proposed hierarchical forecasting and reconciliation framework. We assess forecasting performance before and after applying the neural reconciler, compare classical and learned reconciliation methods, and evaluate robustness under operational disruptions across one-step and multi-step prediction horizons.

\subsection{Data Description and Preprocessing}
\label{sec:data_preprocessing}
 
All experiments are conducted using the smart-card dataset introduced by \cite{nguyen2026multi}, comprising one full year of Rejsekort records (January--December 2017) from a contiguous subnetwork of the Copenhagen S-train system consisting of 12 stations. Passenger demand is observed at a temporal resolution of 20 minutes. Each record captures an individual passenger trip at the origin--destination level, including timestamps, station identifiers, and raw passenger counts. The full OD space comprises $12 \times 12 = 144$ pairs; self-loops are excluded following standard practice, yielding $N_{\mathrm{od}} = 132$ active OD flows. The raw dataset contains 3,784,320 OD-level observations in total.

The 12-station subnetwork represents the common spatial coverage for which station-level demand, OD-level smart-card records, and detailed Banedanmark station-call information could be temporally aligned. The disruption analysis requires not only passenger-demand observations but also operational records on train frequency, delays, and cancellations at compatible spatial and temporal resolutions. Comparable integrated data from additional URT systems were not available to this study. Accordingly, the case study is designed to evaluate forecast reconciliation under operational disruptions within one data-rich network rather than to establish cross-city generalisability or large-network scalability.

To align the data with operational conditions and improve signal quality, we apply a sequence of preprocessing steps informed by \cite{Yang2024AreForecasting}. The dataset is restricted to weekdays and to the operational window between 05:00 and 23:00, removing low-demand overnight periods that would introduce unnecessary noise into the demand signal. A small number of extreme counts are removed by clipping at the 99.95th percentile of the raw passenger count distribution, capping OD demand at 14 passengers per 20-minute interval and affecting fewer than 0.05\% of observations. These extreme values could not be attributed to identifiable operational or external events and are consistent with recording artefacts in the fare collection system rather than genuine demand spikes; retaining them would distort model training without improving representativeness of real operating conditions. After cleaning, the final dataset contains 1,887,600 observations.

Despite temporal aggregation and filtering, the OD data remains highly sparse: approximately 70\% of OD entries are zero across the observation period, reflecting the intrinsic granularity of fine-grained origin--destination demand in URT systems. This sparsity is a defining characteristic of the bottom level of the hierarchy and is a primary reason why estimating the full error covariance matrix $\mathbf{W}$ required by MinT is statistically challenging in this setting, as discussed in Section~\ref{sec:method}. Station-level outflow and inflow series are derived by aggregating OD flows across destinations and origins, respectively, yielding substantially denser signals with zero-demand rates below 12\% and smoother temporal dynamics. These aggregated series form the upper-level inputs to the station-level forecasting model $\mathcal{F}_s$, which is trained to jointly predict both directions, while the OD flows form the bottom-level inputs to $\mathcal{F}_o$.

A structural challenge specific to smart-card data is that departure-indexed
OD flows, and therefore their destination-side inflow aggregates, become
fully observable only as the corresponding trips are completed. At a
real-time prediction cutoff, recent departure cohorts may therefore remain
incomplete, particularly for trips spanning multiple aggregation intervals.
Station outflows, derived directly from tap-ins, do not share this observation
delay. The implications of this asymmetry for forecast coherence are discussed in the
partial-observability remark in Section~3.
 
Both forecasting models take as input lagged demand matrices $\mathbf{X}_t \in \mathbb{R}^{2N \times \tau}$, stacking station outflow and inflow histories, and $\mathbf{Y}_t \in \mathbb{R}^{N_{\mathrm{od}} \times \tau}$, as defined in Section~\ref{sec:method}. Both models additionally incorporate contextual variables including binary indicators for public and school holidays, weather variables (temperature, precipitation, wind speed, and snowfall), and cyclic time encodings for time of day and day of week. For the OD model, additional features capture system reliability conditions at both origin and destination stations over the preceding hour, including train frequency, recorded delays, and cancellation rates, together with OD-pair identifiers to capture pair-specific demand characteristics. Importantly, the feature sets available to $\mathcal{F}_s$ and $\mathcal{F}_o$ are structurally different: the station-level model observes a complete aggregate signal, while the OD model observes sparse, partially incomplete flows augmented with service reliability information unavailable at the station level. This complementarity motivates reconciliation, as enforcing consistency between the two levels allows information learned at one level to reduce systematic errors at the other.
 
The dataset is partitioned into training (70\%), validation (10\%), and test (20\%) sets using a strictly time-based strategy that assigns all observations within a given calendar day to a single partition, preventing information leakage across splits. Numerical features are standardised using statistics computed exclusively from the training set; the same transformation is applied to the validation and test sets to preserve evaluation integrity.

\subsection{Base Forecasting Models}
\label{sec:base_models}
 
We employ two established deep learning architectures as base forecasters, one for each level of the URT demand hierarchy. Both are adopted from prior work and used here as strong, well-validated forecasting backbones rather than as methodological contributions of this paper.
 
For station-level demand, we adopt the Spatial--Dual-Temporal GRU (SDT-GRU) of \cite{Yang2024AreForecasting}, a global sequence-to-sequence model designed for short-term metro ridership forecasting. SDT-GRU combines recurrent modelling with dual-temporal attention to capture both temporal dynamics and system-wide spatial correlations without requiring a manually specified graph structure. We adapt the output layer of SDT-GRU to jointly predict both station outflow and inflow demand. Given the historical input matrix $\mathbf{X}_t \in \mathbb{R}^{2N \times \tau}$, stacking outflow and inflow histories, the model produces joint station-level forecasts $(\widehat{\mathbf{x}}^{\mathrm{out}}_{t+h}, \widehat{\mathbf{x}}^{\mathrm{in}}_{t+h}) = \mathcal{F}_s(\mathbf{X}_t)$ and is trained by minimising mean squared error summed across both outflow and inflow targets.
 
For OD demand, we adopt mGraphSAGE introduced by \cite{nguyen2026multi}, a scalable multi-graph inductive representation learning model in which each OD pair is treated as a node and representations are learned via message passing over multiple relational graphs encoding complementary notions of similarity, including temporal correlation and spatial proximity. This design directly targets the high dimensionality, strong heterogeneity, and extreme sparsity that characterise URT OD demand. Given the historical OD input matrix $\mathbf{Y}_t \in \mathbb{R}^{N_{\mathrm{od}} \times \tau}$ and contextual covariates, mGraphSAGE produces simultaneous forecasts for all OD pairs $\widehat{\mathbf{y}}_{t+h} = \mathcal{F}_o(\mathbf{Y}_t)$ and is trained by minimising mean squared error on OD-level targets.
 
The two models are trained independently on their respective targets. As discussed in Section~\ref{sec:method}, this independence means the resulting base forecasts $\widehat{\mathbf{z}}_{t+h} = (\widehat{\mathbf{x}}^{\mathrm{out}}_{t+h}{}^\top, \widehat{\mathbf{x}}^{\mathrm{in}}_{t+h}{}^\top, \widehat{\mathbf{y}}_{t+h}{}^\top)^\top$ will generally satisfy $\widehat{\mathbf{z}}_{t+h} \notin \mathrm{col}(\mathbf{S})$, violating the conservation constraints that link OD flows to both station outflows and station inflows. The reconciliation methods described in Section~\ref{sec:method} are applied to $\widehat{\mathbf{z}}_{t+h}$ to produce coherent forecasts $\widetilde{\mathbf{z}}_{t+h}$, with the goal of improving predictive accuracy and hierarchical consistency, particularly under disruption conditions.

\subsubsection{Classical Statistical Reconciliation}
 
We compare against four variants of the Minimum Trace (MinT) optimal
combination framework \cite{Wickramasuriya2019OptimalReconciliation}, which
reconciles base forecasts by projecting them onto the coherent subspace
$\mathrm{col}(\mathbf{S})$ while minimising the total variance of
reconciliation errors. All variants share the same reconciliation form:
\[
\widetilde{\mathbf{y}}_{t+h}
=
\mathbf{P}\widehat{\mathbf{z}}_{t+h},
\qquad
\widetilde{\mathbf{z}}_{t+h}
=
\mathbf{S}\widetilde{\mathbf{y}}_{t+h},
\]
where
\[
\mathbf{P}^{\mathrm{MinT}}
=
\bigl(
\mathbf{S}^\top
\mathbf{W}_h^{-1}
\mathbf{S}
\bigr)^{-1}
\mathbf{S}^\top
\mathbf{W}_h^{-1},
\]
and $\mathbf{W}_h$ is the covariance matrix of base forecast errors. The four
variants differ in how $\mathbf{W}_h$ is specified or estimated, spanning a
spectrum from variance-agnostic to fully covariance-aware.

For the one-step evaluation, the covariance estimates used by the
covariance-based MinT variants are computed exclusively from base-forecast
residuals on the training split. Specifically, letting
\[
\mathbf{e}^{\mathrm{train}}_{t,1}
=
\mathbf{z}^{\mathrm{train}}_{t+1}
-
\widehat{\mathbf{z}}^{\mathrm{train}}_{t+1},
\]
the corresponding error statistics are estimated from
$\{\mathbf{e}^{\mathrm{train}}_{t,1}\}_t$. Test-set residuals are not used
in estimating $\mathbf{W}_1$ or any reconciliation projection.

\textbf{MinT-OLS} sets $\mathbf{W}_h \propto \mathbf{I}$, assuming
independent and identically distributed forecast errors. This reduces to a
simple least-squares projection and is computationally efficient, though the
equal-variance assumption is unrealistic in hierarchical settings where
aggregation induces strong cross-series correlations.

\textbf{MinT-WLS} uses a diagonal estimate of $\mathbf{W}_h$ with entries
given by the sample variances of the training-split base forecast residuals,
allowing different scaling across station and OD series while still ignoring
cross-series correlations.

\textbf{MinT-Sample} employs the full sample covariance matrix of the
training-split base forecast residuals, capturing both variances and pairwise
correlations between all series in the hierarchy. Although this provides the
most complete empirical covariance representation, its estimation and
inversion become increasingly demanding and potentially ill-conditioned as
the number of series grows and forecast errors become strongly correlated, a
concern that is particularly acute in the high-dimensional, sparse URT OD
setting.

\textbf{MinT-Shrink} applies a shrinkage estimator to the training-split
sample covariance matrix, regularising it toward its diagonal and improving
numerical stability at the cost of some bias. This provides a practically
robust intermediate between MinT-WLS and MinT-Sample and is generally
recommended when the hierarchy is high-dimensional
\cite{Wickramasuriya2019OptimalReconciliation}.

\subsubsection{Joint Multi-Task Baseline}
\label{sec:joint_baseline}

The reconciliation methods above operate as post-processing steps applied to independently generated base forecasts. An alternative strategy, represented by the multi-task models reviewed in Section~\ref{sec:multilevel_urt}, is to predict station- and OD-level demand jointly within a single model, coupling the two levels through shared representations during training rather than adjusting their outputs afterwards. To assess this alternative empirically on the present case study, we implement a joint multi-task station--OD network.

The model consists of two branches producing per-node embeddings of equal width. The station branch retains the recurrent attention-over-stations cell design of SDT-GRU, but is used as an encoder only: for the one-step target considered here, the decoder and scheduled-sampling machinery of the original architecture reduce to a single application of the recurrent cell and are therefore omitted. The OD branch retains the four-graph \texttt{SAGEConv} structure of mGraphSAGE. The two branches are coupled once, immediately before their output heads, using the aggregation matrices $\mathbf{A}^{\mathrm{out}}$ and $\mathbf{A}^{\mathrm{in}}$ as fixed routing operators: $\mathbf{A}^{\mathrm{out}\top}$ and $\mathbf{A}^{\mathrm{in}\top}$ broadcast each station's embedding to the OD pairs that originate and terminate there, while $\mathbf{A}^{\mathrm{out}}$ and $\mathbf{A}^{\mathrm{in}}$ aggregate OD embeddings back to the corresponding stations. A learned fusion layer in each branch combines its own embedding with the routed context from the other level, after which task-specific heads produce the station outflow and inflow forecasts and the OD flow forecasts.

Crucially, the joint model's forecasts are \emph{not} reconstructed through the summing matrix $\mathbf{S}$. Its two heads produce final predictions directly, so nothing in the architecture constrains $\widehat{\mathbf{x}}_{t+h}$ to equal $\mathbf{A}\widehat{\mathbf{y}}_{t+h}$. The resulting coherence violation is therefore a reported outcome for this baseline rather than a quantity eliminated by construction, which is precisely the distinction the comparison is intended to isolate.

We train two variants of the identical architecture. The first, denoted Joint-NoConsistency, minimises $\mathcal{L} = \mathcal{L}_{\mathrm{station}} + \mathcal{L}_{\mathrm{OD}}$, with both terms computed as mean squared error on the passenger-count scale. It represents multi-task approaches that share representations without penalising aggregation disagreement explicitly. The second, denoted Joint-SoftConsistency, adds a soft consistency penalty,
\[
\mathcal{L} = \mathcal{L}_{\mathrm{station}} + \mathcal{L}_{\mathrm{OD}} + \lambda\,\mathcal{L}_{\mathrm{cons}}, \qquad
\mathcal{L}_{\mathrm{cons}} = \overline{\bigl(\widehat{\mathbf{x}}^{\mathrm{out}}_{t+h} - \mathbf{A}^{\mathrm{out}}\widehat{\mathbf{y}}_{t+h}\bigr)^2} + \overline{\bigl(\widehat{\mathbf{x}}^{\mathrm{in}}_{t+h} - \mathbf{A}^{\mathrm{in}}\widehat{\mathbf{y}}_{t+h}\bigr)^2},
\]
where $\overline{(\cdot)}$ denotes the element-wise mean. This penalises the same structural quantity as the coherence-violation metric of Section~\ref{sec:evaluation_metrics}, but as a training-time regularisation term rather than an exact constraint, and therefore corresponds to the soft aggregation-consistency mechanism used by CATI \cite{yang2026cati}.

This baseline is intended as a controlled, budget-matched representative of
the joint multi-task family rather than as a reimplementation of any specific
published architecture. The model uses the same station and OD encoder designs,
data splits, preprocessing, and a comparable parameter budget as the models in
our framework. This controlled setup enables a direct comparison between joint
multi-task prediction and post-hoc reconciliation under otherwise comparable
experimental conditions.


\subsection{Implementation Details}
\label{sec:implementation}
 
Station-level demand forecasts are generated using the SDT-GRU model of \cite{Yang2024AreForecasting}, adapted to jointly output station outflow and inflow predictions as described in Section~\ref{sec:base_models}. Although the original implementation uses Mean Absolute Error (MAE) as the training objective, preliminary experiments showed that Mean Squared Error (MSE) yields more stable behaviour when station-level predictions are subsequently used as inputs to the reconciliation framework. SDT-GRU is therefore trained with MSE, summed across both outflow and inflow targets, ensuring a consistent optimisation objective across both base forecasters and the reconciler.
 
Input features are selected by progressively augmenting a baseline configuration of lagged inflow--outflow observations and temporal index features with additional groups: holiday indicators, weather variables, and cyclic encodings of time of day and day of week. Each augmented configuration is evaluated against the baseline using paired $t$-tests on validation MSE to assess whether the improvement is statistically significant. Hyperparameters are subsequently tuned over the ranges reported in \cite{Yang2024AreForecasting}. The final configuration uses three recurrent layers with 64 hidden units, four attention heads, and a feed-forward dimension of 64. Training uses the Adam optimiser with a learning rate of $10^{-5}$, batch size 64, and early stopping with patience 30. All experiments use fixed random seeds for reproducibility.
 
OD-level demand is predicted using mGraphSAGE \cite{nguyen2026multi}, trained with a one-step-ahead MSE objective. Optimisation uses Adam with a learning rate of $4 \times 10^{-6}$ and weight decay $5 \times 10^{-4}$. A dropout rate of 0.5 is applied and gradients are clipped to a maximum $\ell_2$ norm of 5. Training uses batch size 64 and early stopping with patience 30. The model follows the architecture reported in \cite{nguyen2026multi} and all experiments use fixed random seeds for reproducibility.
 
The FCR is implemented as a two-layer multilayer perceptron with ReLU activations, taking the concatenated base forecast vector $\widehat{\mathbf{z}}_{t+h} \in \mathbb{R}^M$ as input and producing reconciled bottom-level OD predictions $\widetilde{\mathbf{y}}_{t+h} \in \mathbb{R}^{N_{\mathrm{od}}}$. The full coherent forecast $\widetilde{\mathbf{z}}_{t+h} = \mathbf{S}\widetilde{\mathbf{y}}_{t+h}$ is then recovered via the fixed summing matrix, guaranteeing exact hierarchical coherence by construction regardless of the learned weights.
 
Following \cite{Burba2021Encoder}, the first layer is initialised to approximate the bottom-up mapping, so the untrained reconciler recovers the raw OD base forecasts and provides a stable starting point for optimisation. The reconciler is trained end-to-end using the full-hierarchy MSE loss introduced in Section~\ref{sec:method}, which jointly penalises errors at the station outflow, station inflow, and OD flow levels. Optimisation uses Adam with the same learning rate and early stopping protocol as the base models.

The joint multi-task baseline of Section~\ref{sec:joint_baseline} is trained end-to-end from scratch; no pretrained SDT-GRU or mGraphSAGE weights are used. Both branches use a hidden width of 80 units, selected so that the total parameter count of 766{,}643 falls within $\pm 20\%$ of the combined parameter count of the components it is compared against (SDT-GRU, 614{,}402; mGraphSAGE, 149{,}249; FCR, 38{,}280; total 801{,}931). Matching the budget in this way ensures that differences in the reported results are not attributable to a substantially larger or smaller model. Optimisation uses Adam with a learning rate of $10^{-3}$, batch size 64, and early stopping with patience 30. A higher learning rate than that used for the individual base forecasters is appropriate here because the joint model is trained from random initialisation rather than fine-tuned. All runs use fixed random seeds for reproducibility, and the training data, temporal split, and standardisation statistics are identical to those used for the base forecasters.

The consistency weight $\lambda$ of the Joint-SoftConsistency variant is selected over $\{0.01, 0.1, 1.0\}$ using validation full-hierarchy MSE, computed with the same metric implementation used throughout the evaluation. The selected value is $\lambda = 0.01$. Results for the full sweep are reported in Section~\ref{sec:joint_results}, as they illustrate the accuracy--coherence trade-off induced by the penalty.

\subsection{Multi-Step Reconciliation and Disruption Evaluation}
\label{sec:multistep_setup}

\paragraph{Multi-step forecasting}

To evaluate performance beyond the one-step-ahead setting, we extend the pipeline to multi-step prediction over $H=6$ successive horizons. At each time $t$, both base models produce forecasts for $\{t+1,\ldots,t+H\}$.
 
The SDT-GRU station-level model supports multi-step forecasting natively through its recurrent decoder. To enable multi-horizon OD forecasting, the final projection layer of mGraphSAGE is replaced with a layer that maps each OD-node embedding to an $H$-dimensional output. The complete OD model is then retrained end-to-end using an $H$-step MSE loss, such that the GraphSAGE layers and the multi-step projection head are optimised jointly.
 
For each horizon $\ell$, the station-level forecast contains both outflow and inflow predictions,
\[
\widehat{\mathbf{x}}_{t+\ell}
=
\begin{bmatrix}
\widehat{\mathbf{x}}^{\mathrm{out}}_{t+\ell} \\
\widehat{\mathbf{x}}^{\mathrm{in}}_{t+\ell}
\end{bmatrix}
\in\mathbb{R}^{2N},
\]
while $\widehat{\mathbf{y}}_{t+\ell}\in\mathbb{R}^{N(N-1)}$ denotes the corresponding directed OD-flow forecast. The two base models therefore produce the horizon-wise forecast sequences
\[
\left(
\widehat{\mathbf{x}}_{t+1},\ldots,
\widehat{\mathbf{x}}_{t+H}
\right)
\quad\text{and}\quad
\left(
\widehat{\mathbf{y}}_{t+1},\ldots,
\widehat{\mathbf{y}}_{t+H}
\right).
\]
 
\paragraph{Shared-weight reconciler}
 
Multi-step neural reconciliation is performed using a shared extension of FCR, denoted S-FCR. At each horizon $\ell\in\{1,\ldots,H\}$, the concatenated base forecast
\[
\widehat{\mathbf{z}}_{t+\ell}
=
\begin{bmatrix}
\widehat{\mathbf{x}}_{t+\ell} \\
\widehat{\mathbf{y}}_{t+\ell}
\end{bmatrix}
\]
is passed through the same reconciler parameters:
\[
\widetilde{\mathbf{y}}_{t+\ell}
=
\mathbf{P}_{\theta}
\left(
\widehat{\mathbf{z}}_{t+\ell}
\right),
\qquad
\widetilde{\mathbf{z}}_{t+\ell}
=
\mathbf{S}\widetilde{\mathbf{y}}_{t+\ell},
\qquad
\ell=1,\ldots,H.
\]
Exact hierarchical coherence is guaranteed at every horizon through the fixed summing matrix $\mathbf{S}$. The reconciler is trained by minimising the average full-hierarchy MSE across all horizons:
\[
\mathcal{L}(\theta)
=
\frac{1}{H}
\sum_{\ell=1}^{H}
\left\|
\mathbf{z}_{t+\ell}
-
\widetilde{\mathbf{z}}_{t+\ell}
\right\|_2^2.
\]
Sharing the reconciler parameters across horizons limits parameter growth and allows a common nonlinear reconciliation mapping to be learned over the complete prediction window.

\paragraph{Horizon-specific MinT-Sample}
MinT-Sample is used as the classical baseline in the multi-step evaluation. Its covariance matrix is estimated separately for each prediction horizon. After training the station-level and OD-level forecasting models, the trained models are applied to the training split without additional parameter updates to obtain multi-step base forecasts. For each horizon $\ell$, the corresponding training residuals are computed as
\[
\mathbf{e}^{\mathrm{train}}_{t,\ell}
=
\mathbf{z}^{\mathrm{train}}_{t+\ell}
-
\widehat{\mathbf{z}}^{\mathrm{train}}_{t+\ell},
\]
and the sample covariance matrix is estimated as
\[
\mathbf{W}_{\ell}
=
\operatorname{Cov}
\left(
\mathbf{e}^{\mathrm{train}}_{t,\ell}
\right).
\]
The horizon-specific MinT projection is then given by
\[
\mathbf{P}^{\mathrm{MinT}}_{\ell}
=
\left(
\mathbf{S}^{\top}
\mathbf{W}_{\ell}^{-1}
\mathbf{S}
\right)^{-1}
\mathbf{S}^{\top}
\mathbf{W}_{\ell}^{-1},
\]
with reconciled forecasts obtained as
\[
\widetilde{\mathbf{y}}^{\mathrm{MinT}}_{t+\ell}
=
\mathbf{P}^{\mathrm{MinT}}_{\ell}
\widehat{\mathbf{z}}_{t+\ell},
\qquad
\widetilde{\mathbf{z}}^{\mathrm{MinT}}_{t+\ell}
=
\mathbf{S}
\widetilde{\mathbf{y}}^{\mathrm{MinT}}_{t+\ell}.
\]
Each covariance and projection matrix is estimated using residuals from its corresponding horizon and is applied only to test forecasts at that same horizon. Test-set residuals are not used in covariance estimation.

\paragraph{Disruption evaluation}
In addition to aggregate test performance, we conduct a targeted evaluation under operationally atypical conditions to assess model robustness. For the one-step setting, we consider three categories of atypical conditions: operational disruptions caused by train delays and cancellations, adverse weather, and public holidays. For the multi-step setting, we focus on operational disruptions because they can produce abrupt changes in passenger demand and increase disagreement between forecasts generated at different aggregation levels.
 
Disrupted forecasting intervals are identified using the supply-side reliability indicators included in the OD-model feature set (Section~\ref{sec:data_preprocessing}), following the threshold-based criteria of \cite{nguyen2026multi}. A 20-minute interval is classified as disrupted when the cancellation rate is non-zero or when the mean delay at the origin or destination station exceeds a specified threshold. We consider cumulative delay thresholds of 60, 180, and 300 seconds, representing increasing levels of disruption severity. Intervals with no recorded cancellations and zero mean delay constitute the undisrupted reference condition.

Because the forecasting dataset stores the reliability indicators after aggregation into 20-minute intervals, it does not preserve the number of individual delayed or cancelled train calls underlying each interval. To characterise the frequency of operational disruptions more directly, we therefore reprocessed the raw Banedanmark RDS records at station-call granularity. We applied the same station-renaming, train-line assignment, entry-call filtering, and cancellation-identification procedures used when constructing the forecasting dataset, and restricted the analysis to the same 12 stations considered in the forecasting experiments.
 
Among 1,600,522 valid station-call records, 122,020 station calls experienced delays exceeding 60 seconds, 42,495 exceeded 180 seconds, and 24,412 exceeded 300 seconds. In addition, 40,264 station-call records were identified as cancellations. Table~\ref{tab:disruption_frequency} reports these frequencies and their shares of all valid station calls. The delay thresholds are cumulative rather than mutually exclusive; thus, a station call delayed by more than 300 seconds is also included in the 60- and 180-second categories. These values represent station-call observations rather than unique operational incidents, because consistent incident identifiers are not available in the source data. The temporal distribution of these observations over the study period is presented in \ref{app:disruption_distribution}, Figure~\ref{fig:daily_disruption}. The disruption-performance tables reported later in the paper use OD-level forecast observations as their evaluation samples and should therefore not be interpreted as counts of independent disruptions.

\begin{table}[t]
\centering
\caption{Frequency of operational disruption observations in the raw Banedanmark records for the 12-station study network. Counts refer to individual station-call records, and delay thresholds are cumulative.}
\label{tab:disruption_frequency}
\begin{tabular}{lrr}
\toprule
Operational condition &
Station-call records &
Share of all station calls \\
\midrule
Delay $>60$ s  & 122,020 & 7.62\% \\
Delay $>180$ s & 42,495  & 2.66\% \\
Delay $>300$ s & 24,412  & 1.53\% \\
Cancellation  & 40,264   & 2.52\% \\
\midrule
Total station calls & 1,600,522 & 100\% \\
\bottomrule
\end{tabular}
\end{table}

\subsection{Evaluation Metrics}
\label{sec:evaluation_metrics}

We evaluate the forecasting methods along two distinct dimensions: predictive accuracy and structural coherence. Let $\bar{\mathbf{y}}^{(m)}_{t+h}\in\mathbb{R}^{N_{\mathrm{od}}}$ and $\bar{\mathbf{x}}^{(m)}_{t+h}\in\mathbb{R}^{2N}$ denote, respectively, the OD-level forecast and the stacked station outflow--inflow forecast produced by method $m$ for horizon $h$, with $N_{\mathrm{od}} = N(N-1)$. For the unreconciled base forecasts, $\bar{\mathbf{y}}^{(m)}_{t+h}=\widehat{\mathbf{y}}_{t+h}$ and $\bar{\mathbf{x}}^{(m)}_{t+h}=(\widehat{\mathbf{x}}^{\mathrm{out}}_{t+h}{}^\top, \widehat{\mathbf{x}}^{\mathrm{in}}_{t+h}{}^\top)^\top$. For a reconciliation method, they correspond to $\widetilde{\mathbf{y}}^{(m)}_{t+h}$ and $\widetilde{\mathbf{x}}^{(m)}_{t+h}$.
 
The full-hierarchy observed and predicted vectors are defined as
\begin{equation}
\mathbf{z}_{t+h}
=
\begin{bmatrix}
\mathbf{x}^{\mathrm{out}}_{t+h} \\
\mathbf{x}^{\mathrm{in}}_{t+h} \\
\mathbf{y}_{t+h}
\end{bmatrix}
\in \mathbb{R}^M,
\qquad
\bar{\mathbf{z}}^{(m)}_{t+h}
=
\begin{bmatrix}
\bar{\mathbf{x}}^{(m)}_{t+h} \\
\bar{\mathbf{y}}^{(m)}_{t+h}
\end{bmatrix}
\in \mathbb{R}^M,
\label{eq:full_hierarchy_vectors}
\end{equation}
with $M = 2N + N(N-1)$ as defined in Section~\ref{sec:method}. All metrics are computed after transforming the forecasts back to the original passenger-count scale and over the same $T$ evaluation intervals.
 
\begin{table}[H]
\centering
\footnotesize
\caption{Evaluation metrics. Accuracy metrics compare forecasts with observed demand, whereas the coherence-violation metric evaluates whether station- and OD-level forecasts satisfy the aggregation constraints.}
\begin{tabular}{@{}llp{6.2cm}@{}}
\toprule
\textbf{Dimension}
&
\textbf{Metric}
&
\textbf{Description}
\\
\midrule
 
OD accuracy
&
OD MSE
&
Mean squared error between
$\mathbf{y}_{t+h}$ and $\bar{\mathbf{y}}^{(m)}_{t+h}$
\\
 
OD accuracy
&
OD MAE
&
Mean absolute error between
$\mathbf{y}_{t+h}$ and $\bar{\mathbf{y}}^{(m)}_{t+h}$
\\
 
OD accuracy
&
OD WMAPE
&
Total absolute OD forecasting error relative to total observed OD demand
\\
 
OD accuracy
&
OD SMAPE
&
Mean symmetric percentage error across individual OD observations
\\
 
\midrule
 
Station accuracy
&
Station MSE
&
Mean squared error between the stacked outflow--inflow vector $(\mathbf{x}^{\mathrm{out}}_{t+h}{}^\top, \mathbf{x}^{\mathrm{in}}_{t+h}{}^\top)^\top$ and $\bar{\mathbf{x}}^{(m)}_{t+h}$
\\
 
Station accuracy
&
Station MAE
&
Mean absolute error between the stacked outflow--inflow vector and $\bar{\mathbf{x}}^{(m)}_{t+h}$
\\
 
\midrule
 
Full-hierarchy accuracy
&
Full MSE
&
Mean squared error between
$\mathbf{z}_{t+h}$ and $\bar{\mathbf{z}}^{(m)}_{t+h}$
\\
 
Full-hierarchy accuracy
&
Full MAE
&
Mean absolute error between
$\mathbf{z}_{t+h}$ and $\bar{\mathbf{z}}^{(m)}_{t+h}$
\\
 
\midrule
 
Structural coherence
&
Coherence violation MSE
&
Mean squared discrepancy, summed over outflow and inflow, between the station forecast and the station demand implied by aggregation of the corresponding OD forecast
\\
 
\bottomrule
\end{tabular}
\label{tab:evaluation_metrics}
\end{table}
 
For a generic error vector $\mathbf{e}_{t,h}\in\mathbb{R}^{n}$, MSE and MAE are computed as
\begin{align}
\mathrm{MSE}
&=
\frac{1}{Tn}
\sum_{t=1}^{T}
\left\|\mathbf{e}_{t,h}\right\|_2^2,
\label{eq:evaluation_mse}
\\
\mathrm{MAE}
&=
\frac{1}{Tn}
\sum_{t=1}^{T}
\left\|\mathbf{e}_{t,h}\right\|_1,
\label{eq:evaluation_mae}
\end{align}
where $n$ is set to $N_{\mathrm{od}}$ for OD-level evaluation, $2N$ for station-level evaluation, and $M$ for full-hierarchy evaluation, with the error vector defined analogously in each case using the corresponding observed and predicted vectors; for OD-level evaluation, $\mathbf{e}_{t,h} = \mathbf{y}_{t+h}-\bar{\mathbf{y}}^{(m)}_{t+h}$.
 
Because OD demand is sparse and varies substantially across station pairs and time periods, we additionally report Weighted Mean Absolute Percentage Error (WMAPE) and Symmetric Mean Absolute Percentage Error (SMAPE):
\begin{align}
\mathrm{WMAPE}
&=
100
\frac{
\displaystyle
\sum_{t=1}^{T}\sum_{j=1}^{N_{\mathrm{od}}}
\left|
y_{t+h,j}
-
\bar{y}^{(m)}_{t+h,j}
\right|
}{
\displaystyle
\sum_{t=1}^{T}\sum_{j=1}^{N_{\mathrm{od}}}
\left|y_{t+h,j}\right|
},
\label{eq:wmape}
\\
\mathrm{SMAPE}
&=
\frac{100}{TN_{\mathrm{od}}}
\sum_{t=1}^{T}\sum_{j=1}^{N_{\mathrm{od}}}
\frac{
2\left|
y_{t+h,j}
-
\bar{y}^{(m)}_{t+h,j}
\right|
}{
\left|y_{t+h,j}\right|
+
\left|\bar{y}^{(m)}_{t+h,j}\right|
}.
\label{eq:smape}
\end{align}
For SMAPE, the contribution is defined as zero when both the observed and predicted values are zero. WMAPE expresses the total absolute forecasting error relative to total passenger demand and is therefore directly interpretable at the network level. SMAPE measures relative error at the individual OD-observation level and provides additional information about performance on lower-demand OD pairs and time periods. The percentage-based metrics are reported together with MSE and MAE because WMAPE remains weighted toward high-volume flows, while SMAPE can be sensitive when both observed and predicted demand are close to zero.
 
Structural coherence is evaluated independently of observed demand. The coherence-violation MSE for method $m$ is defined as
\begin{equation}
\mathrm{CV}^{(m)}_{h}
=
\frac{1}{2TN}
\sum_{t=1}^{T}
\left(
\left\|
\bar{\mathbf{x}}^{\mathrm{out},(m)}_{t+h}
-
\mathbf{A}^{\mathrm{out}}\bar{\mathbf{y}}^{(m)}_{t+h}
\right\|_2^2
+
\left\|
\bar{\mathbf{x}}^{\mathrm{in},(m)}_{t+h}
-
\mathbf{A}^{\mathrm{in}}\bar{\mathbf{y}}^{(m)}_{t+h}
\right\|_2^2
\right).
\label{eq:coherence_violation}
\end{equation}
A value of zero indicates that the station- and OD-level forecasts satisfy both aggregation constraints of equation~\eqref{eq:conservation} exactly. For the independently generated base forecasts, this quantity measures their cross-level structural disagreement, jointly across outflow and inflow. The interval-level form,
\begin{equation}
D^{(m)}_{t,h}
=
\frac{1}{2N}
\left(
\left\|
\bar{\mathbf{x}}^{\mathrm{out},(m)}_{t+h}
-
\mathbf{A}^{\mathrm{out}}\bar{\mathbf{y}}^{(m)}_{t+h}
\right\|_2^2
+
\left\|
\bar{\mathbf{x}}^{\mathrm{in},(m)}_{t+h}
-
\mathbf{A}^{\mathrm{in}}\bar{\mathbf{y}}^{(m)}_{t+h}
\right\|_2^2
\right),
\label{eq:interval_disagreement}
\end{equation}
is used later to analyse how forecast disagreement varies across operating conditions.
 
For a coherent reconciled forecast,
\begin{equation}
\widetilde{\mathbf{x}}^{\mathrm{out},(m)}_{t+h}
=
\mathbf{A}^{\mathrm{out}}\widetilde{\mathbf{y}}^{(m)}_{t+h},
\qquad
\widetilde{\mathbf{x}}^{\mathrm{in},(m)}_{t+h}
=
\mathbf{A}^{\mathrm{in}}\widetilde{\mathbf{y}}^{(m)}_{t+h},
\label{eq:reconciled_coherence}
\end{equation}
and its coherence violation is therefore zero up to numerical precision. Structural coherence remains distinct from forecasting accuracy: a forecast may satisfy both aggregation constraints while still differing from the observed passenger demand.

\section{Results}

\subsection{One-Step Reconciliation}

\subsubsection{Base Forecast Accuracy}

We first evaluate the standalone predictive performance of the two base forecasting models prior to any reconciliation. This establishes the quality of the inputs to the reconciliation stage and quantifies the degree of incoherence that reconciliation must correct.

\paragraph{Station-level forecasting}
 
Table~\ref{tab:model_eval} reports station outflow forecasting performance. To ensure a like-for-like comparison with the baseline models, all figures in this table correspond to station outflow only; SDT-GRU's inflow forecasts, which the model produces jointly as described in Section~\ref{sec:base_models}, are reported separately as part of the joint outflow--inflow station metric used in the reconciliation results (Table~\ref{tab:single_step_hierarchy_performance}). SDT-GRU achieves the lowest MAE (2.1187) and MSE (9.5857), outperforming all competing methods including PVCGN \cite{Liu2020PhysicalVirtual}, a widely used graph convolutional benchmark for metro passenger flow prediction, with the improvement statistically significant at the 1\% level. The machine learning baselines (Random Forest and XGBoost) and the Historical Average (HA) exhibit substantially higher errors, reflecting their limited capacity to capture the complex spatio-temporal dynamics of station-level demand. These simpler models produce mean errors closer to zero, indicating predictions concentrated near the global mean, but this reduced bias comes at the cost of higher MAE and MSE across the full demand distribution.

\begin{table}[H]
\centering
\caption{Station outflow forecasting performance. Bold: best; underline: second best. $^{**}$ denotes significance at the 1\% level versus the second-best model.}
\begin{tabular}{@{}lccc@{}}
\toprule
\textbf{Model} & \textbf{MAE} & \textbf{MSE} & \textbf{ME} \\
\midrule
\textbf{SDT-GRU} & \textbf{2.1187}$^{**}$ & \textbf{9.5857} & $-$0.2988 \\
\underline{PVCGN} & \underline{2.2022} & \underline{10.3874} & $-$0.3699 \\
RF & 2.6294 & 14.3785 & 0.0447 \\
XGBoost & 2.5893 & 13.9714 & 0.0370 \\
HA & 2.4965 & 16.1234 & 0.1180 \\
\bottomrule
\end{tabular}
\label{tab:model_eval}

\vspace{0.3em}
\parbox{0.95\linewidth}{\footnotesize
Figures correspond to station outflow forecasts only. PVCGN, RF, XGBoost, and HA are single-output models and are compared on outflow prediction; SDT-GRU's jointly predicted inflow forecasts, and the combined outflow--inflow station metric, are reported in Table~\ref{tab:single_step_hierarchy_performance}.}
\end{table}

\paragraph{OD-level forecasting}

Table~\ref{tab:od_model_comparison} compares the OD forecasting models using absolute, percentage-based, and bias metrics. mGraphSAGE achieves the lowest MSE, MAE, and WMAPE, with an MSE of 0.7823 that is significantly lower than that of the second-best model at the 1\% level. Its consistent performance across these metrics indicates that it controls both large forecasting errors and aggregate absolute error more effectively than the alternative models. XGBoost provides the closest competition, ranking second in MSE, MAE, and WMAPE, while also exhibiting the smallest absolute mean error.
 
The SMAPE ranking differs, with the Historical Average and RF obtaining lower values. This reflects the sensitivity of SMAPE to the large proportion of zero and low-demand OD observations: methods that frequently reproduce zero or small values can perform favourably under this metric even when they are less effective at capturing larger demand variations. Accordingly, both methods have substantially higher MSE and WMAPE than mGraphSAGE. Moreover, the Historical Average result is calculated only for observations with previously available same-slot history, excluding 36,300 cold-start observations.
 
Overall, mGraphSAGE provides the strongest and most consistent performance across the principal absolute and aggregate relative-error measures. These results support its use as the OD-level base forecaster in the subsequent reconciliation experiments, where accurate bottom-level predictions are required before information from the station level is incorporated.
 
\begin{table}[H]
\centering
\small
\caption{OD-level forecasting performance. WMAPE and SMAPE are reported as percentages. Bold and underlined values indicate the best and second-best results, respectively. For ME, values closest to zero indicate lower systematic bias. $^{**}$ denotes significance at the 1\% level for MSE relative to the second-best model. Lower is better for all metrics except that ME is evaluated by its absolute magnitude.}
\label{tab:od_model_comparison}
\begin{tabular}{@{}lccccc@{}}
\toprule
\textbf{Model}
& \textbf{MSE}
& \textbf{MAE}
& \textbf{WMAPE (\%)}
& \textbf{SMAPE (\%)}
& \textbf{ME} \\
\midrule
mGraphSAGE
& \textbf{0.7823}$^{**}$
& \textbf{0.4999}
& \textbf{82.3562}
& 162.8879
& 0.0131 \\
 
XGBoost
& \underline{0.8494}
& \underline{0.5291}
& \underline{87.1792}
& 162.8265
& 0.0050 \\
 
RF
& 0.9123
& 0.5375
& 88.5562
& \underline{144.5887}
& 0.0091 \\
 
Historical Average
& 1.0918
& 0.5519
& 91.6412
& \textbf{91.7491}
& 0.0111 \\
\bottomrule
\end{tabular}
 
\vspace{0.3em}
\parbox{0.95\linewidth}{\footnotesize
The Historical Average is evaluated on 340,164 of 376,464 OD observations; 36,300 cold-start observations without prior same-slot history are excluded.}
\end{table}

\paragraph{Incoherence of base forecasts}
Across both hierarchical levels, the deep learning models establish strong
predictive baselines for the subsequent reconciliation stage. However,
because SDT-GRU and mGraphSAGE are trained independently on separate
objectives, their outputs do not enforce the aggregation constraints linking
OD flows to station outflows and inflows. The base forecast vector
$\widehat{\mathbf{z}}_{t+h}
=
\left(
\widehat{\mathbf{x}}^{\mathrm{out}\top}_{t+h},
\widehat{\mathbf{x}}^{\mathrm{in}\top}_{t+h},
\widehat{\mathbf{y}}^\top_{t+h}
\right)^\top
$
therefore generally satisfies
$\widehat{\mathbf{z}}_{t+h}\notin\operatorname{col}(\mathbf{S})$.

\subsubsection{One-Step Reconciliation Performance}

Tables~\ref{tab:single_step_od_performance} and \ref{tab:single_step_hierarchy_performance} report one-step forecasting performance over the same 2,852 test intervals. The first table evaluates OD-level accuracy using absolute and percentage-based error metrics, whereas the second reports station-level and full-hierarchy accuracy together with structural coherence violation. Lower values indicate better performance for all metrics.
 
\begin{table}[H]
\centering
\caption{One-step OD-level forecasting performance over 2,852 evaluation intervals. WMAPE and SMAPE are reported as percentages. Bold and underlined values indicate the best and second-best results, respectively. Lower is better for all metrics.}
\label{tab:single_step_od_performance}
\begin{tabular}{lcccc}
\toprule
\textbf{Method}
& \textbf{OD MSE}
& \textbf{OD MAE}
& \textbf{OD WMAPE (\%)}
& \textbf{OD SMAPE (\%)} \\
\midrule
Base forecast
& 0.7823
& 0.4999
& 82.3562
& 162.8873 \\
 
RF reconciler
& 0.7758
& 0.4948
& 81.5231
& \textbf{147.4048} \\
 
MinT-OLS
& 0.7796
& 0.4985
& 82.1290
& 162.8969 \\
 
MinT-WLS
& 0.7774
& 0.4973
& 81.9262
& 162.9080 \\
 
MinT-Sample
& \textbf{0.7645}
& \textbf{0.4924}
& \textbf{81.1319}
& 162.7866 \\
 
MinT-Shrink
& 0.7684
& \underline{0.4937}
& \underline{81.3352}
& 162.7859 \\
 
FCR
& \underline{0.7663}
& 0.4946
& 81.4937
& \underline{162.2648} \\
\bottomrule
\end{tabular}
\end{table}
 
\begin{table}[H]
\centering
\caption{One-step station-level and full-hierarchy forecasting performance and structural coherence. Bold and underlined values indicate the best and second-best forecasting accuracy, respectively. Coherence-violation MSE values below $10^{-10}$ are reported as $\approx 0$. Lower is better for all metrics.}
\label{tab:single_step_hierarchy_performance}
\resizebox{\textwidth}{!}{%
\begin{tabular}{lccccc}
\toprule
\textbf{Method}
& \textbf{Station MSE}
& \textbf{Station MAE}
& \textbf{Full MSE}
& \textbf{Full MAE}
& \textbf{Coherence violation MSE} \\
\midrule
Base forecast
& 9.3639
& 2.1181
& 2.1026
& 0.7488
& 0.6501 \\
 
RF reconciler
& 9.4248
& 2.1319
& 2.1064
& 0.7467
& $\approx 0$ \\
 
MinT-OLS
& 9.3250
& 2.1137
& 2.0943
& 0.7470
& $\approx 0$ \\
 
MinT-WLS
& 9.2741
& 2.1071
& 2.0846
& 0.7449
& $\approx 0$ \\
 
MinT-Sample
& \textbf{9.1114}
& \textbf{2.0945}
& \textbf{2.0487}
& \textbf{0.7389}
& $\approx 0$ \\
 
MinT-Shrink
& 9.1614
& \underline{2.0971}
& 2.0596
& \underline{0.7403}
& $\approx 0$ \\
 
FCR
& \underline{9.1549}
& 2.1058
& \underline{2.0568}
& 0.7425
& $\approx 0$ \\
\bottomrule
\end{tabular}%
}
\end{table}
 
The independently generated base forecasts have a coherence-violation MSE of 0.6501, indicating disagreement between the direct station-level forecast and the station outflow and inflow implied by the OD forecast. By contrast, all reconciliation methods satisfy both aggregation constraints up to numerical precision. The largest coherence-violation MSE among the reconciled methods is $1.802\times10^{-13}$ for FCR; the MinT values are of the order of $10^{-29}$, and the RF value is numerically zero. These differences arise from floating-point computation and do not represent meaningful differences in structural coherence.
 
Because all reconciliation methods produce coherent forecasts, their relative performance is assessed through forecasting accuracy. MinT-Sample achieves the lowest error in seven of the eight reported accuracy metrics. Relative to the unreconciled base forecasts, it reduces OD MSE from 0.7823 to 0.7645, station MSE from 9.3639 to 9.1114, and full-hierarchy MSE from 2.1026 to 2.0487, corresponding to reductions of 2.28\%, 2.70\%, and 2.56\%, respectively. Its OD WMAPE decreases from 82.36\% to 81.13\%. MinT-Shrink achieves the second-lowest OD, station, and full-hierarchy MAE and the second-lowest OD WMAPE.
 
FCR remains closely competitive with the strongest MinT variants. It obtains the second-lowest MSE at the OD, station, and full-hierarchy levels, reducing these errors relative to the base forecasts by 2.05\%, 2.23\%, and 2.18\%, respectively. FCR also achieves the second-lowest SMAPE while satisfying the aggregation constraint by construction and without requiring explicit estimation of a forecast-error covariance matrix.
 
The RF reconciler exhibits a different performance profile. It achieves the lowest SMAPE, decreasing it from 162.89\% to 147.40\%, and also improves OD MSE, MAE, and WMAPE. However, it increases station MSE and MAE and slightly increases full-hierarchy MSE. Its improvement in relative OD error therefore does not translate into uniform accuracy gains across the hierarchy. Similarly, MinT-OLS and MinT-WLS improve most absolute-error metrics but do not improve SMAPE relative to the base forecasts.
 
The percentage-based metrics provide a complementary view of performance. Because all methods are evaluated using the same observed OD demand, WMAPE is proportional to OD MAE and consequently produces the same method ranking, while expressing the error relative to total observed passenger demand. SMAPE instead evaluates relative error at the individual OD-observation level and gives greater influence to observations with small demand values. The high SMAPE values reflect the sparsity of the OD demand data and the sensitivity of percentage errors when observed and predicted flows are close to zero. Together, the results show that MinT-Sample provides the strongest overall one-step accuracy, FCR remains closely competitive across the hierarchy, and RF achieves the lowest observation-level relative error measured by SMAPE.

\subsubsection{Oracle Analysis of Reconciliation Potential}
 
To examine how the quality of the station-level signal affects reconciliation, we conduct an oracle experiment in which the predicted station input to the trained FCR is replaced by the observed station outflow and inflow demand at the target interval. The OD input remains the original mGraphSAGE base forecast, and the FCR parameters are unchanged. This counterfactual setting isolates the additional forecasting gains that can be obtained when accurate station-level information, for both outflow and inflow, is available. All settings are evaluated on the same 2,852 test intervals.
 
Tables~\ref{tab:oracle_od_performance} and \ref{tab:oracle_hierarchy_performance} compare the oracle result with the unreconciled base forecasts, FCR using the real station forecast, and MinT-Sample, which achieves the strongest overall accuracy under the standard one-step setting.
 
\begin{table}[H]
\centering
\caption{OD-level forecasting performance with real and oracle station-level inputs. WMAPE and SMAPE are reported as percentages. The oracle FCR uses observed station outflow and inflow demand as its station-level input while retaining the original OD base forecast. Lower is better for all metrics.}
\label{tab:oracle_od_performance}
\resizebox{\textwidth}{!}{%
\begin{tabular}{lcccc}
\toprule
\textbf{Setting}
& \textbf{OD MSE}
& \textbf{OD MAE}
& \textbf{OD WMAPE (\%)}
& \textbf{OD SMAPE (\%)} \\
\midrule
Base forecast
& 0.7823
& 0.4999
& 82.3562
& 162.8873 \\
 
FCR with real station input
& 0.7663
& 0.4946
& 81.4937
& 162.2648 \\
 
MinT-Sample
& 0.7645
& 0.4924
& 81.1319
& 162.7866 \\
\midrule
 
FCR with oracle station input
& 0.5130
& 0.4181
& 68.8794
& 159.7866 \\
\bottomrule
\end{tabular}%
}
\end{table}
 
\begin{table}[H]
\centering
\caption{Station-level and full-hierarchy forecasting performance with real and oracle station-level inputs. Coherence-violation MSE values below $10^{-10}$ are reported as $\approx 0$. Lower is better for all metrics.}
\label{tab:oracle_hierarchy_performance}
\resizebox{\textwidth}{!}{%
\begin{tabular}{lccccc}
\toprule
\textbf{Setting}
& \textbf{Station MSE}
& \textbf{Station MAE}
& \textbf{Full MSE}
& \textbf{Full MAE}
& \textbf{Coherence violation MSE} \\
\midrule
Base forecast
& 9.3639
& 2.1181
& 2.1026
& 0.7488
& 0.6501 \\
 
FCR with real station input
& 9.1549
& 2.1058
& 2.0568
& 0.7425
& $\approx 0$ \\
 
MinT-Sample
& 9.1114
& 2.0945
& 2.0487
& 0.7389
& $\approx 0$ \\
\midrule
 
FCR with oracle station input
& 0.0077
& 0.0519
& 0.4353
& 0.3617
& $\approx 0$ \\
\bottomrule
\end{tabular}%
}
\end{table}
 
Replacing the predicted station input with observed station outflow and inflow demand substantially improves OD forecasting accuracy. Relative to the unreconciled base forecast, the oracle FCR reduces OD MSE from 0.7823 to 0.5130, corresponding to a 34.42\% reduction. OD MAE and WMAPE decrease by 16.36\%, from 0.4999 to 0.4181 and from 82.36\% to 68.88\%, respectively. Compared with FCR using the real station forecast, the oracle input reduces OD MSE by 33.05\% and OD MAE by 15.48\%. It also substantially outperforms MinT-Sample, reducing OD MSE by 32.89\%.
 
The reduction in SMAPE is smaller, from 162.89\% for the base forecast to 159.79\% under the oracle setting. This reflects the sensitivity of SMAPE to the many zero and low-demand OD observations: improved prediction of aggregate flows and demand peaks has a much stronger effect on MSE and WMAPE than on the observation-level symmetric percentage error.
 
The oracle station input yields a station MSE of 0.0077 and a full-hierarchy MSE of 0.4353, showing that the reconciled forecasts closely recover the observed station outflow and inflow totals while also substantially improving accuracy across the hierarchy. The small remaining station error is an accuracy error rather than a structural-coherence violation: it measures the difference between the observed station outflow and inflow demand and the station forecast obtained by aggregating the reconciled OD predictions through $\mathbf{A}^{\mathrm{out}}$ and $\mathbf{A}^{\mathrm{in}}$. Although the observed station demand is supplied as an input, the trained FCR is not explicitly constrained to reproduce this input exactly. Instead, it uses the station signal jointly with the OD base forecasts to generate coherent OD predictions.
 
Both the real-input and oracle-input FCR forecasts have coherence-violation MSE values below $2\times10^{-13}$, confirming that they satisfy both aggregation constraints up to numerical precision. The oracle information therefore improves forecasting accuracy rather than structural coherence, which is already guaranteed by the reconciliation architecture regardless of input quality.
 
Overall, the oracle analysis shows that the quality of the station-level input,
across both outflow and inflow, materially affects the accuracy gains obtainable
through reconciliation. The substantially larger improvement obtained in the
oracle setting illustrates the value of more accurate cross-level information.
However, this result should be interpreted as a counterfactual diagnostic rather
than as fully attainable forecasting headroom: observed target-interval station
demand is not available at the prediction cutoff, and destination-side inflow
aggregates in particular become complete only as the corresponding tap-outs are
observed. The oracle gain therefore reflects both the value of improved
station-level forecasting and information that is unavailable in real time.

\subsubsection{Analysis of Reconciliation Behaviour}

To complement the aggregate metrics reported in Tables~\ref{tab:single_step_od_performance} and \ref{tab:single_step_hierarchy_performance}, we examine error distributions, OD-pair-level error patterns, and representative forecast trajectories. These analyses clarify how reconciliation modifies the base forecasts, where the improvements originate spatially, and where limitations remain.

\begin{figure}[H]
    \centering
    \makebox[\textwidth][c]{%
        \includegraphics[width=\textwidth]{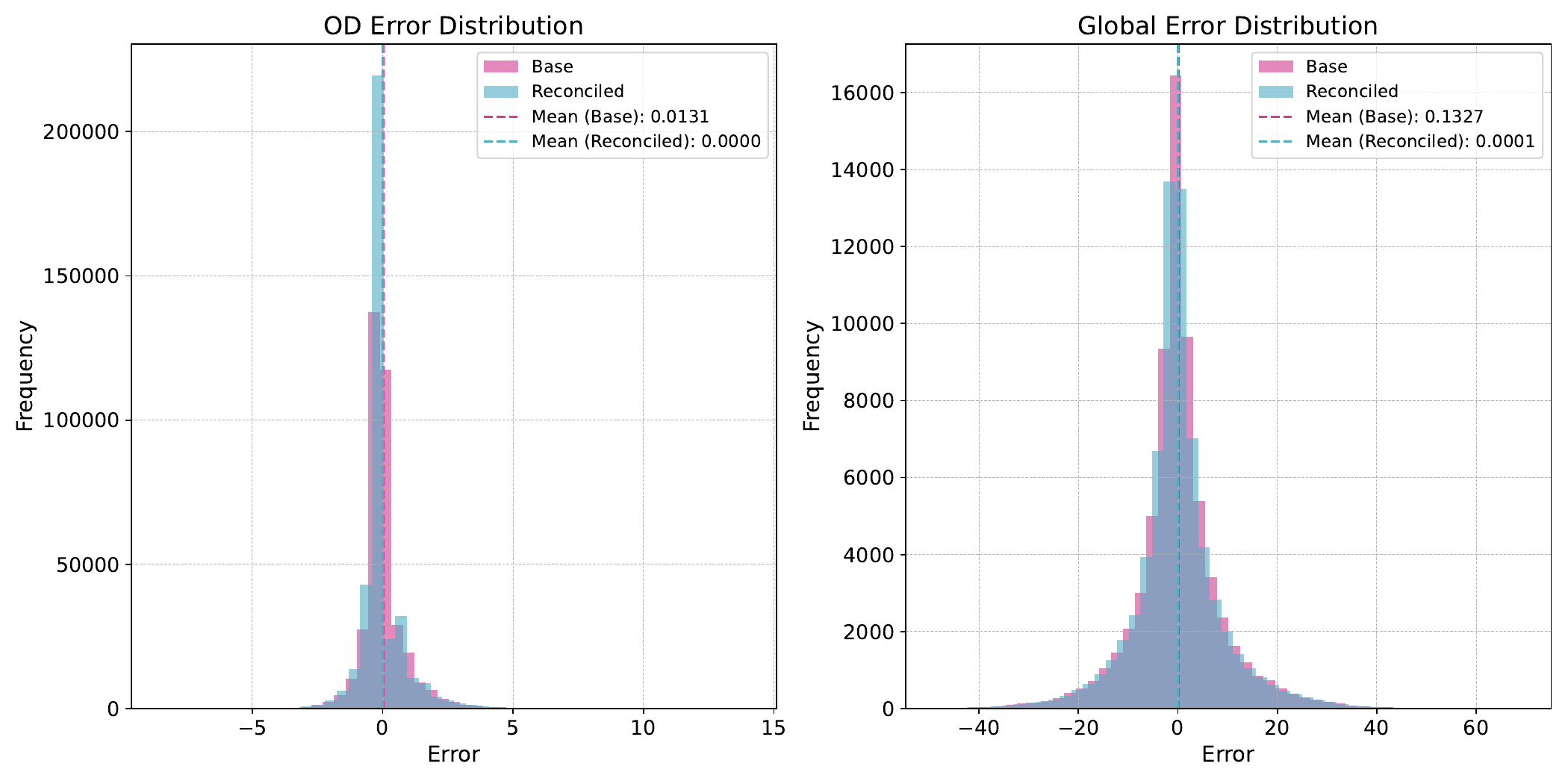}%
    }
    \caption{Error distributions for base and reconciled forecasts at the OD and station levels.}
    \label{fig:error_dist}
\end{figure}

\paragraph{Error distributions}
Figure~\ref{fig:error_dist} shows the forecast error distributions at both the OD and station levels before and after reconciliation. In both cases the distributions are approximately centred around zero with bell-shaped profiles, though with sharper peaks and heavier tails than a Gaussian. Reconciliation produces a visibly tighter concentration around zero, particularly at the OD level, indicating reduced forecast dispersion. It also eliminates the positive bias present in the base forecasts: the mean error falls from 0.0131 to approximately zero at the OD level and from 0.1327 to approximately zero at the station level. This confirms that reconciliation improves not only aggregate accuracy but also systematic bias across the hierarchy.

\begin{figure}[H]
    \centering
    \makebox[\textwidth][c]{%
        \includegraphics[width=\textwidth]{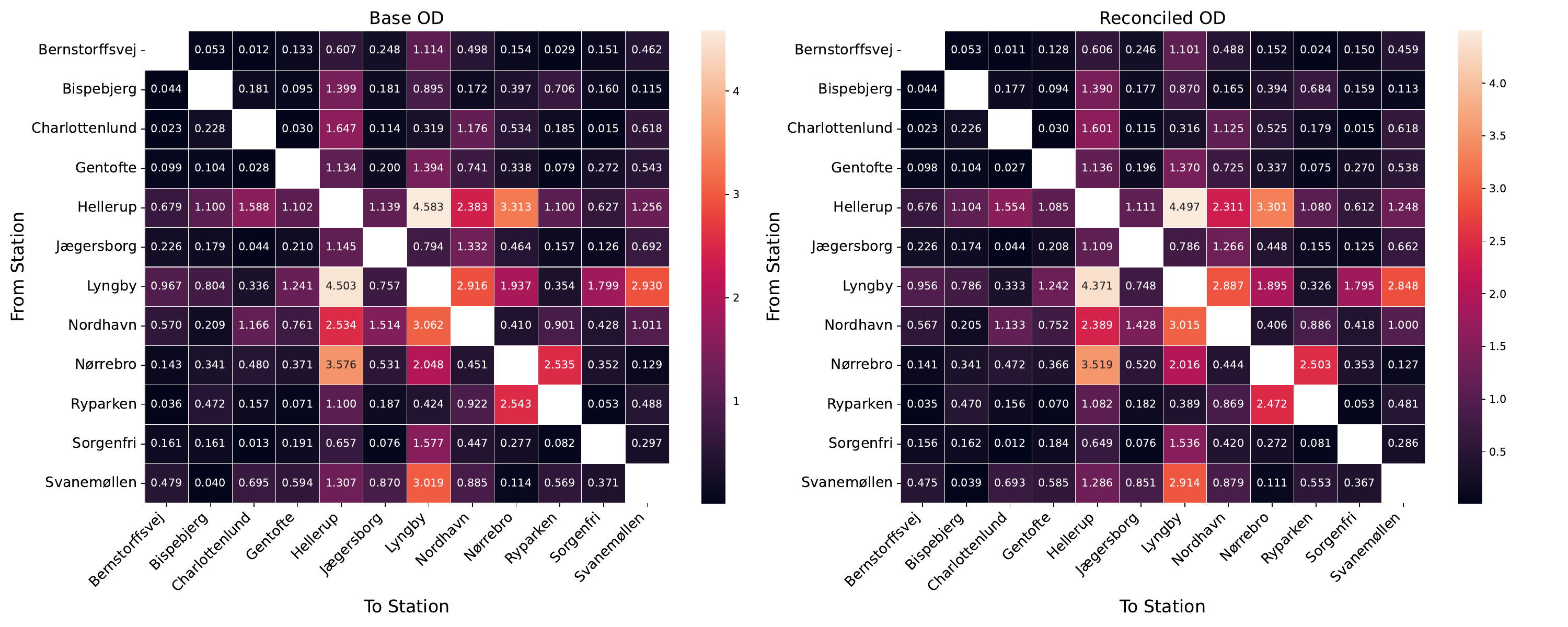}%
    }
    \caption{OD-pair MSE before and after reconciliation.}
    \label{fig:heatmap_odpairs}
\end{figure}

\paragraph{Spatial distribution of errors}
Figure~\ref{fig:heatmap_odpairs} visualises the OD-pair MSE before and after reconciliation. Errors are concentrated in high-demand OD pairs involving major interchange stations such as Lyngby, Hellerup, Nordhavn, and Nørrebro, where larger passenger volumes produce greater absolute variability and therefore a harder forecasting problem. OD pairs involving smaller peripheral stations such as Bernstorffsvej and Sorgenfri exhibit near-zero errors, largely because their demand remains close to zero across most intervals. Reconciliation produces the largest absolute improvements precisely in the high-demand OD pairs: for example, the MSE for the Lyngby$\rightarrow$Hellerup pair decreases from 4.503 to 4.371, with similar reductions visible for other high-frequency pairs involving Nordhavn and Hellerup. Improvements for small-volume pairs remain marginal, as the limited demand variability leaves little room for correction. This pattern confirms that reconciliation is most beneficial where cross-level information is richest, namely in the high-demand regions of the network where station-level and OD-level signals are most informative about each other.

\begin{figure}[H]
    \centering
    \makebox[\textwidth][c]{%
        \includegraphics[width=1.1\textwidth]{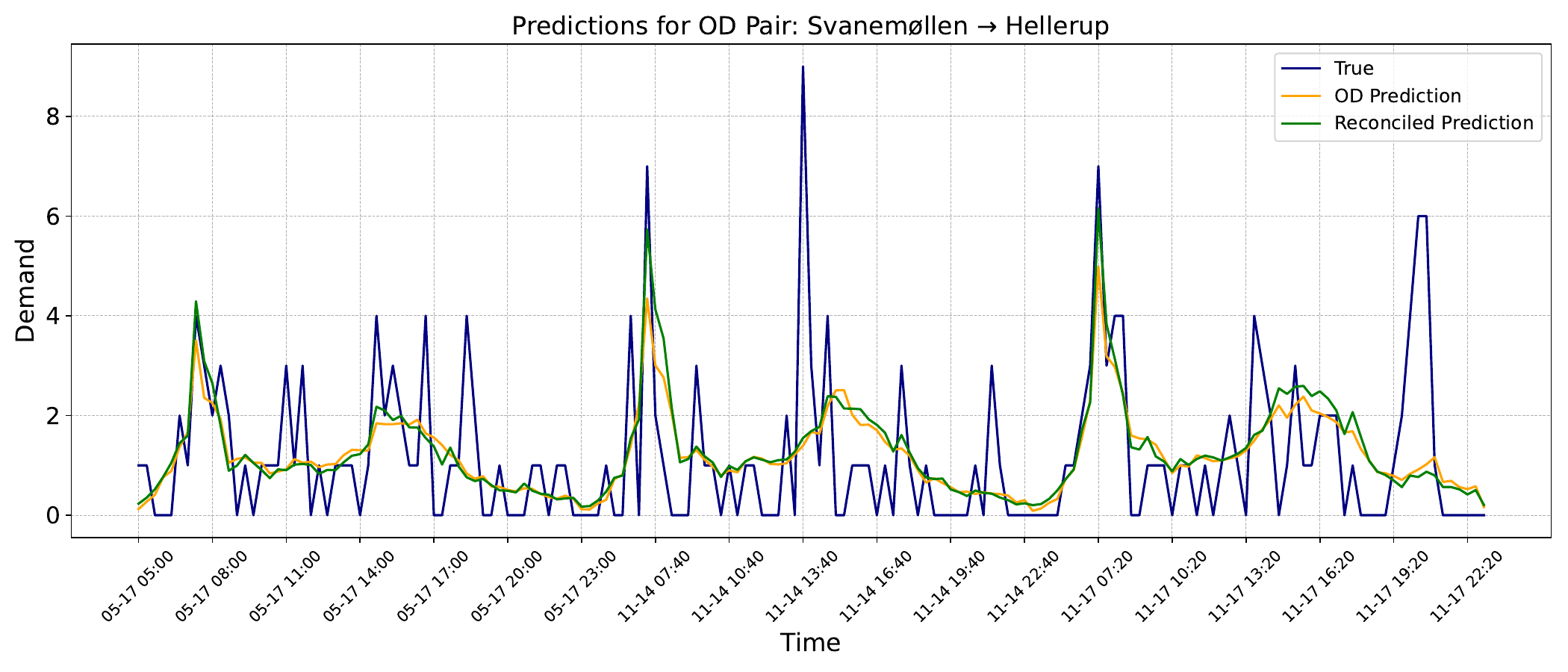}%
    }
    \caption{Base and reconciled OD forecasts for the Svanemøllen$\rightarrow$Hellerup OD pair.}
    \label{fig:preds_od}
\end{figure}

\paragraph{OD-level forecast trajectories}
Figure~\ref{fig:preds_od} shows representative OD-level forecasts for the Svanemøllen $\rightarrow$ Hellerup pair across several test days. The reconciled forecasts closely track the base mGraphSAGE predictions, confirming that reconciliation acts as a targeted corrective adjustment rather than overriding the temporal dynamics learned by the base model. Visible improvements are concentrated around peak demand periods, where reconciled forecasts more accurately capture the magnitude of demand spikes. Both models nonetheless struggle with extreme isolated peaks that are weakly correlated with recent observations — a limitation inherent to one-step forecasting models that rely on lagged demand inputs, and particularly pronounced at the OD level where many series are sparse and large observations infrequent.

\begin{figure}[H]
    \centering
    \makebox[\textwidth][c]{%
        \includegraphics[width=1.1\textwidth]{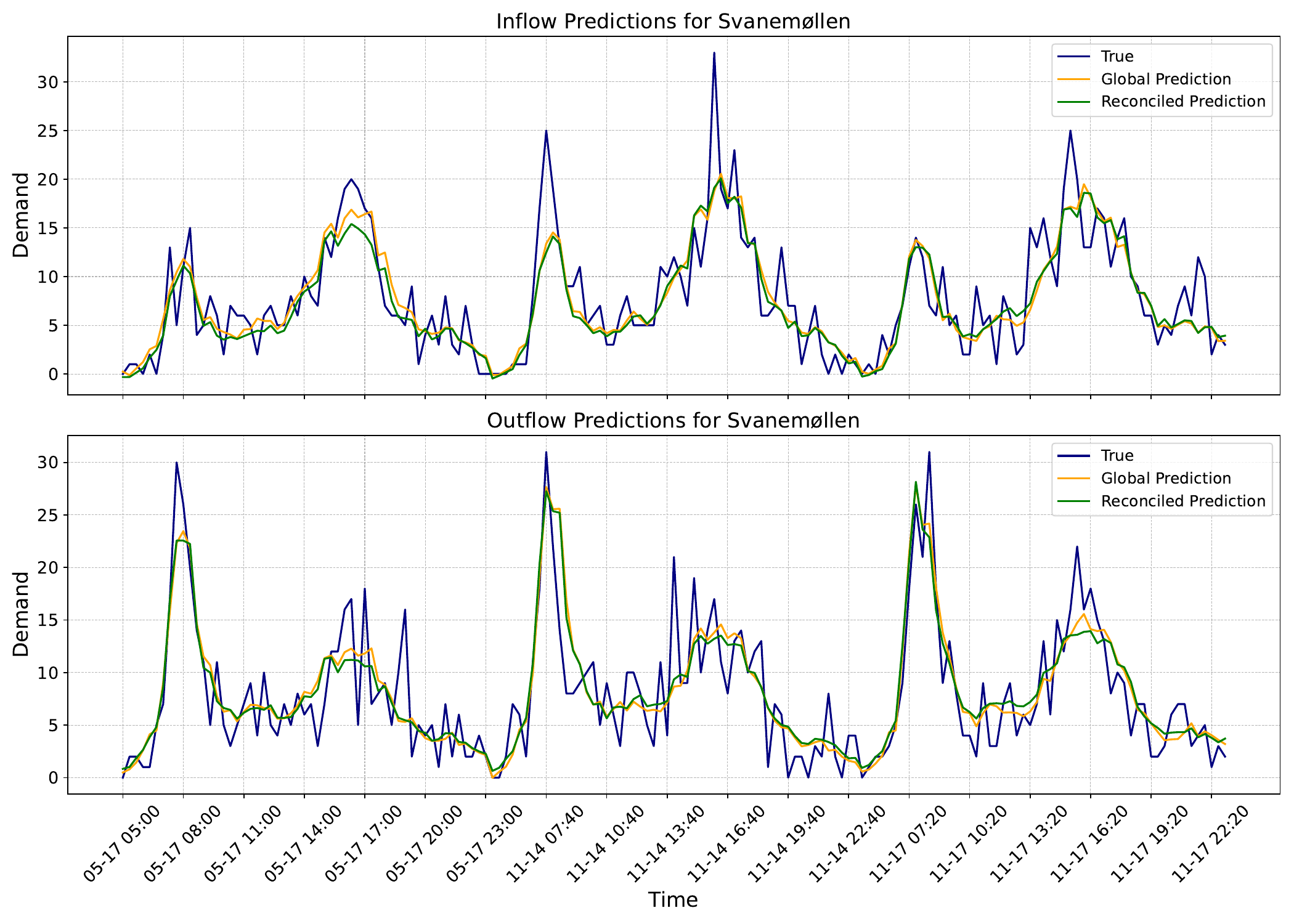}%
    }
    \caption{Base and reconciled station-level inflow and outflow forecasts for Svanemøllen station.}
    \label{fig:preds_station}
\end{figure}

\paragraph{Station-level forecast trajectories}
Figure~\ref{fig:preds_station} presents analogous results for inflow and outflow forecasting at Svanemøllen station. Station-level predictions are substantially smoother than their OD-level counterparts, reflecting the noise-reducing effect of spatial aggregation. Recurring peak patterns, particularly morning commuting demand around 07:00, are accurately captured by both SDT-GRU and the reconciled forecasts, confirming that the station-level model successfully learns regular temporal demand structures. The reconciled station-level forecasts remain close to the SDT-GRU predictions, with small corrections visible around several peak periods. These adjustments account for the improved station-level accuracy metrics reported earlier while preserving the overall temporal structure of the base model.

Taken together, these analyses confirm a consistent picture of reconciliation behaviour: it preserves the large-scale temporal patterns captured by the base models, eliminates systematic bias, and concentrates improvements in the high-demand regions of the network where cross-level information is most useful for correction.

\subsubsection{Comparison with a Joint Multi-Task Baseline}
\label{sec:joint_results}

The preceding results evaluate reconciliation as a post-processing step applied to independently generated base forecasts. We now compare this strategy against the joint multi-task baseline described in Section~\ref{sec:joint_baseline}, in which station- and OD-level demand are predicted by a single model that couples the two levels during encoding rather than adjusting their outputs afterwards. All results are computed on the same 2,852 test intervals and with the same metric implementations used throughout this section.

Table~\ref{tab:lambda_sweep} reports the consistency-weight sweep for the
Joint-SoftConsistency variant on the validation set. Increasing $\lambda$
produces a monotonic reduction in coherence violation, from 1.1952 at
$\lambda=0$ to 0.0543 at $\lambda=1.0$, while full-hierarchy accuracy varies
non-monotonically across the tested values. The penalty therefore behaves as
intended by progressively moving the two prediction heads towards agreement.
The value $\lambda=0.01$ attains the lowest validation full-hierarchy MSE and
is used in the comparison that follows, although its advantage over
$\lambda=0$ is marginal (2.3661 against 2.3672).

\begin{table}[H]
\centering
\caption{Consistency-weight sweep for the Joint-SoftConsistency variant, evaluated on the validation set. $\lambda=0$ corresponds to the Joint-NoConsistency variant. The selected value is $\lambda=0.01$. Lower is better for all metrics.}
\label{tab:lambda_sweep}
\begin{tabular}{lcccc}
\toprule
$\boldsymbol{\lambda}$
& \textbf{OD MSE}
& \textbf{Station MSE}
& \textbf{Full MSE}
& \textbf{Coherence violation MSE} \\
\midrule
0 (no consistency) & 0.8853 & 10.5172 & 2.3672 & 1.1952 \\
0.01 (selected)    & 0.8811 & 10.5337 & \textbf{2.3661} & 0.4463 \\
0.1                & 0.9093 & 10.5131 & 2.3868 & 0.2932 \\
1.0                & 0.9261 & 10.5740 & 2.4104 & \textbf{0.0543} \\
\bottomrule
\end{tabular}
\end{table}

Table~\ref{tab:joint_comparison} reports test-set performance for the two joint variants alongside the unreconciled base forecasts, MinT-Sample, and FCR.

\begin{table}[H]
\centering
\caption{One-step comparison of the joint multi-task baseline with the unreconciled base forecasts and the two strongest reconciliation methods, over the same 2,852 evaluation intervals. WMAPE and SMAPE are reported as percentages. Coherence-violation MSE values below $10^{-10}$ are reported as $\approx 0$. Lower is better for all metrics.}
\label{tab:joint_comparison}
\resizebox{\textwidth}{!}{%
\begin{tabular}{lccccccc}
\toprule
\textbf{Method}
& \textbf{OD MSE}
& \textbf{OD MAE}
& \textbf{OD WMAPE (\%)}
& \textbf{OD SMAPE (\%)}
& \textbf{Station MSE}
& \textbf{Full MSE}
& \textbf{Coherence violation MSE} \\
\midrule
Base forecast & 0.7823 & 0.4999 & 82.3562 & 162.8873 & 9.3639 & 2.1026 & 0.6501 \\
MinT-Sample & \textbf{0.7645} & \textbf{0.4924} & \textbf{81.1319} & 162.7866 & \textbf{9.1114} & \textbf{2.0487} & $\approx 0$ \\
FCR & \underline{0.7663} & \underline{0.4946} & \underline{81.4937} & \textbf{162.2648} & \underline{9.1549} & \underline{2.0568} & $\approx 0$ \\
\midrule
Joint-NoConsistency & 0.8186 & 0.4957 & 81.6751 & 165.0247 & 9.4587 & 2.1478 & 1.2297 \\
Joint-SoftConsistency ($\lambda=0.01$) & 0.8133 & 0.5182 & 85.3816 & 163.1981 & 9.4709 & 2.1453 & 0.4551 \\
\bottomrule
\end{tabular}%
}
\end{table}

\paragraph{Structural coherence}
The coherence results separate the two strategies clearly. Both reconciliation methods reconstruct their forecasts through the summing matrix and therefore satisfy the aggregation constraints up to numerical precision. The joint model does not. Trained without a consistency term, its coherence violation of 1.2297 is larger than that of the independently generated base forecasts (0.6501): coupling the two branches through a shared representation did not, by itself, induce any structural agreement between their outputs. Adding the soft consistency penalty at the selected weight reduces the violation
to 0.4551, approximately 30\% below the base forecasts, confirming that the
penalty exerts real pressure towards agreement. However, the penalised model
does not achieve exact coherence. Even the most aggressive setting examined,
$\lambda = 1.0$, reaches a coherence violation of 0.0543 on the validation set.
By contrast, FCR and MinT-Sample are coherent by construction, satisfying the
aggregation constraints up to numerical precision. Thus, a soft consistency
penalty can reduce cross-level disagreement, but it does not provide the exact
structural guarantee obtained through reconciliation.

\paragraph{Forecasting accuracy}
The joint baseline does not improve accuracy relative to the independently trained base forecasters. Both variants record higher OD MSE (0.8186 and 0.8133) than the base forecasts (0.7823), and consequently higher than FCR (0.7663) and MinT-Sample (0.7645). The same ordering holds at the station and full-hierarchy levels. Training the two levels jointly, under a parameter budget matched to the combined size of the models it replaces, therefore did not recover the accuracy obtained by optimising each level separately and reconciling afterwards on this dataset. This accuracy comparison should be interpreted with more caution than the coherence comparison. The joint architecture was sized to the parameter budget rather than tuned for predictive performance, and its learning rate, coupling design, and training schedule were not subjected to a search comparable to that performed for the base forecasters. All configurations were also trained once, without repeated-seed variance
estimates. The result reported here therefore indicates that this particular budget-matched, untuned joint design underperforms post-hoc reconciliation on this dataset, rather than establishing a general accuracy ordering between joint learning and reconciliation. The coherence finding, by contrast, follows from the architectures themselves: a model whose outputs are not reconstructed through $\mathbf{S}$ has no mechanism guaranteeing exact coherence, irrespective of tuning.

The comparison supports the positioning of this work set out in Section~\ref{sec:related}. Joint multi-task learning and forecast reconciliation address the same structural relationship by different means, and the soft consistency mechanism used by recent joint models does reduce cross-level disagreement. Only reconstruction through the summing matrix, however, removes that disagreement entirely, and in this case study it does so while also achieving better forecasting accuracy.

\subsection{Multi-Step Forecasting and Reconciliation}

We extend the evaluation to six-step forecasting, corresponding to prediction horizons of up to two hours. The SDT-GRU station-level model supports multi-step prediction through its recurrent decoder. The mGraphSAGE OD model is modified by extending its final projection layer to produce six outputs per OD embedding, as described in Section~\ref{sec:multistep_setup}. The S-FCR is applied at each horizon using a common set of parameters across all forecast steps.

Because the multi-step OD forecaster is retrained using the six-step objective,
the \(T+1\) forecasts in this experiment are not directly comparable with the
dedicated one-step results in Section~5.1.2; all percentage reductions reported
below are computed relative to the corresponding multi-step base forecasts.
 
Tables~\ref{tab:multistep_mse_comparison} and \ref{tab:multistep_percent_comparison} report the per-horizon OD, station, and full-hierarchy MSE and their reductions relative to the corresponding unreconciled base forecasts. MinT-Sample is selected as the strongest classical reconciliation baseline from the one-step evaluation.
 
\begin{table}[H]
\centering
\small
\caption{Per-horizon forecasting MSE for the unreconciled base forecasts, S-FCR, and MinT-Sample. Bold values indicate the best reconciled result at each horizon. Lower is better.}
\begin{tabular}{@{}llcccccc@{}}
\toprule
\textbf{Metric} &
\textbf{Method} &
\textbf{T+1} &
\textbf{T+2} &
\textbf{T+3} &
\textbf{T+4} &
\textbf{T+5} &
\textbf{T+6} \\
\midrule
\multirow{3}{*}{OD MSE}
& Base forecast
    & 0.8059 & 0.8058 & 0.8048 & 0.8039 & 0.8053 & 0.8109 \\
& S-FCR
    & \textbf{0.7731} & \textbf{0.7735} & \textbf{0.7739}
    & \textbf{0.7745} & \textbf{0.7758} & \textbf{0.7785} \\
& MinT-Sample
    & 0.7844 & 0.7841 & 0.7832 & 0.7838 & 0.7858 & 0.7897 \\
\midrule
\multirow{3}{*}{Station MSE}
& Base forecast
    & 9.4893 & 9.6082 & 9.6814 & 9.7884 & 9.8986 & 10.0275 \\
& S-FCR
    & \textbf{9.2627} & \textbf{9.3477} & \textbf{9.3893}
    & \textbf{9.4329} & \textbf{9.5107} & \textbf{9.6260} \\
& MinT-Sample
    & 9.4023 & 9.5179 & 9.5573 & 9.6237 & 9.7060 & 9.8373 \\
\midrule
\multirow{3}{*}{Full-hierarchy MSE}
& Base forecast
    & 2.1418 & 2.1600 & 2.1704 & 2.1861 & 2.2043 & 2.2288 \\
& S-FCR
    & \textbf{2.0792} & \textbf{2.0926} & \textbf{2.0993}
    & \textbf{2.1066} & \textbf{2.1196} & \textbf{2.1397} \\
& MinT-Sample
    & 2.1102 & 2.1278 & 2.1331 & 2.1438 & 2.1581 & 2.1816 \\
\bottomrule
\end{tabular}
\label{tab:multistep_mse_comparison}
\end{table}
 
\begin{table}[H]
\centering
\small
\caption{Percentage reduction in MSE relative to the corresponding unreconciled base forecasts across prediction horizons. Bold values indicate the larger reduction at each horizon.}
\begin{tabular}{@{}llcccccc@{}}
\toprule
\textbf{Metric} &
\textbf{Method} &
\textbf{T+1} &
\textbf{T+2} &
\textbf{T+3} &
\textbf{T+4} &
\textbf{T+5} &
\textbf{T+6} \\
\midrule
\multirow{2}{*}{OD MSE}
& S-FCR
    & \textbf{4.07} & \textbf{4.01} & \textbf{3.84}
    & \textbf{3.66} & \textbf{3.66} & \textbf{4.00} \\
& MinT-Sample
    & 2.67 & 2.69 & 2.68 & 2.50 & 2.42 & 2.61 \\
\midrule
\multirow{2}{*}{Station MSE}
& S-FCR
    & \textbf{2.39} & \textbf{2.71} & \textbf{3.02}
    & \textbf{3.63} & \textbf{3.92} & \textbf{4.00} \\
& MinT-Sample
    & 0.92 & 0.94 & 1.28 & 1.68 & 1.95 & 1.90 \\
\midrule
\multirow{2}{*}{Full-hierarchy MSE}
& S-FCR
    & \textbf{2.92} & \textbf{3.12} & \textbf{3.28}
    & \textbf{3.64} & \textbf{3.84} & \textbf{4.00} \\
& MinT-Sample
    & 1.47 & 1.49 & 1.72 & 1.94 & 2.09 & 2.12 \\
\bottomrule
\end{tabular}
\label{tab:multistep_percent_comparison}
\end{table}
 
Forecasting accuracy remains broadly stable across the six horizons, with only moderate variation in the MSE values. Both S-FCR and MinT-Sample improve upon the corresponding unreconciled base forecasts at every horizon and for all three reported accuracy measures.
 
S-FCR reduces OD MSE by 3.66\%--4.07\% across the prediction window, whereas MinT-Sample achieves reductions of 2.42\%--2.69\%. The improvements obtained by both methods remain broadly stable across horizons, with S-FCR producing the lower OD MSE at every examined horizon.
 
A similar pattern is observed at the station level. S-FCR reduces station MSE by 2.39\% at T+1, with the reduction increasing to 4.00\% at T+6. MinT-Sample reduces station MSE by 0.92\%--1.95\% across the same horizons. These results indicate that the cross-level adjustment provided by S-FCR improves the accuracy of the station forecasts obtained from the reconciled OD predictions throughout the multi-step prediction window.
 
For the full hierarchy, S-FCR achieves MSE reductions of 2.92\%--4.00\%, compared with 1.47\%--2.12\% for MinT-Sample. Both reconciliation methods satisfy the station--OD aggregation constraints up to numerical precision at every horizon. Their differences therefore concern forecasting accuracy rather than structural coherence.
 
Overall, the multi-step results show that reconciliation remains beneficial throughout the six-step prediction window. S-FCR achieves consistently larger reductions in OD, station, and full-hierarchy MSE than MinT-Sample, despite using one shared set of reconciler parameters across all horizons. The shared-weight design therefore achieves stable multi-step forecasting improvements without requiring separately learned reconciler parameters for each prediction horizon.

\subsection{Reconciliation under Operational Disruption Conditions}
\label{sec:disruption_results}

Operational disruptions can alter passenger-demand patterns and affect the station-level and OD-level forecasting models differently. As a result, the independently generated forecasts may exhibit larger structural disagreement, increasing the potential value of reconciliation for improving predictive accuracy. We therefore evaluate whether reconciliation remains effective under train delays and cancellations and examine how disruption conditions are related to demand characteristics, base-forecast coherence violations, and reconciliation gains.
 
The main analysis considers train cancellations and delays exceeding 60 seconds. These conditions are operationally meaningful while providing sufficient observations for comparison with regular operating conditions. The forecasting and reconciliation models are trained on the naturally occurring mixture of regular and disrupted observations, without disruption-specific oversampling, reweighting, or fine-tuning. The results therefore assess the same forecasting pipeline across different operating conditions. Disruption variables are derived from Banedanmark station-call records and aligned with the corresponding 20-minute passenger-demand intervals.
 
\subsubsection{One-Step Reconciliation under Disruption}
 
In Tables~\ref{tab:cancellation_disruptions} and \ref{tab:representative_delay_disruptions}, ``OD observations'' denotes the number of OD-pair forecasts satisfying each condition and included in the error calculation. Multiple OD observations may correspond to the same 20-minute interval because all concurrently active OD pairs are evaluated.
 
\paragraph{Train cancellations}
Table~\ref{tab:cancellation_disruptions} compares the base forecasts, FCR, and MinT-Sample under cancellation-free and cancellation-affected conditions. In the absence of cancellations, both reconciliation methods reduce OD forecasting error by approximately 2\%. Under origin-side cancellations, FCR reduces OD MSE by 2.35\%, compared with 1.20\% for MinT-Sample. The difference is larger under destination-side cancellations, where FCR reduces OD MSE by 5.56\%, compared with 3.64\% for MinT-Sample. These results show that reconciliation remains beneficial when train
cancellations occur, with FCR yielding the larger point reduction in both
cancellation-affected subsets.
 
\begin{table}[H]
\centering
\caption{One-step OD forecasting performance under train cancellation conditions. Reductions are measured relative to the unreconciled base OD forecast.}
\resizebox{\textwidth}{!}{
\begin{tabular}{lcccccc}
\toprule
\textbf{Condition} &
\textbf{Base OD MSE} &
\textbf{FCR OD MSE} &
\textbf{MinT OD MSE} &
\textbf{FCR reduction (\%)} &
\textbf{MinT reduction (\%)} &
\textbf{OD observations}\\
\midrule
Origin cancellations $=0$
    & 0.7828 & 0.7667 & 0.7649 & 2.05 & 2.28 & 373,144 \\
Origin cancellations $>0$
    & 0.7285 & 0.7114 & 0.7198 & 2.35 & 1.20 & 3,320 \\
\midrule
Destination cancellations $=0$
    & 0.7808 & 0.7650 & 0.7631 & 2.02 & 2.26 & 373,285 \\
Destination cancellations $>0$
    & 0.9622 & 0.9087 & 0.9272 & 5.56 & 3.64 & 3,179 \\
\bottomrule
\end{tabular}%
}
\label{tab:cancellation_disruptions}
\end{table}
 
\paragraph{Operational delays}
Table~\ref{tab:representative_delay_disruptions} reports results for the representative delay threshold of 60 seconds. Under origin-side delays, the base OD MSE increases from 0.7980 in delay-free observations to 1.0158 when the mean delay exceeds 60 seconds. FCR reduces the disrupted-condition error by 5.21\%, compared with 4.33\% for MinT-Sample. Under destination-side delays, FCR reduces OD MSE by 6.09\%, compared with 4.97\% for MinT-Sample. Thus, both methods remain effective under the representative delay condition,
with FCR yielding larger point reductions in the affected subsets.
 
\begin{table}[H]
\centering
\caption{One-step OD forecasting performance under representative delay conditions.
Reductions are measured relative to the unreconciled base OD forecast. The reported
delay categories are not exhaustive: observations with mean delay in $(0,60]$ s are
excluded from both rows.}
\resizebox{\textwidth}{!}{
\begin{tabular}{lcccccc}
\toprule
\textbf{Condition} &
\textbf{Base OD MSE} &
\textbf{FCR OD MSE} &
\textbf{MinT OD MSE} &
\textbf{FCR reduction (\%)} &
\textbf{MinT reduction (\%)} &
\textbf{OD observations}\\
\midrule
Origin delay $=0$ s
    & 0.7980 & 0.7819 & 0.7803 & 2.02 & 2.22 & 340,182 \\
Origin delay $>60$ s
    & 1.0158 & 0.9629 & 0.9718 & 5.21 & 4.33 & 3,208 \\
\midrule
Destination delay $=0$ s
    & 0.7915 & 0.7763 & 0.7747 & 1.92 & 2.13 & 340,580 \\
Destination delay $>60$ s
    & 0.9072 & 0.8520 & 0.8621 & 6.09 & 4.97 & 3,074 \\
\bottomrule
\end{tabular}}
\label{tab:representative_delay_disruptions}
\end{table}
 
Overall, the one-step results show that reconciliation improves OD forecasting
accuracy under representative cancellation and delay conditions. FCR yields
larger point reductions than MinT-Sample in each disruption-affected subset,
whereas MinT-Sample yields slightly larger reductions in the corresponding
unaffected subsets. The variation in reconciliation gains across conditions
motivates the subsequent analysis of demand-pattern shifts and base-forecast
coherence violations.
 
Results for the less frequent 180- and 300-second delay thresholds are reported in \ref{app:additional_disruption_results} as supplementary sensitivity analyses. Holiday- and weather-related results are also included there to examine reconciliation under other atypical operating conditions.
 
\subsubsection{Demand-Pattern Shift under Operational Disruptions}
\label{sec:demand_shift}

To examine whether operational disruptions correspond to a different demand regime, we compare the empirical passenger-demand distributions in undisrupted and disrupted 20-minute intervals. This diagnostic uses the same weekday and 05:00--23:00 temporal population as the forecasting experiments. Of the 14,300 modelling intervals, 14,285 contain valid station-call observations and can therefore be assigned a network-level disruption status. An interval is classified as disrupted when the network-wide mean delay across the valid station calls in that interval exceeds 60 seconds. This criterion is more selective than classifying an interval based on a single delayed call, which would identify a large proportion of busy operating periods as disrupted.

Among the 14,285 evaluable intervals, 1,241 are classified as disrupted and 13,044 as undisrupted. Table~\ref{tab:disruption_demand_shift} shows that the disrupted intervals exhibit higher and more variable passenger demand. Mean OD demand increases from 0.6093 to 0.7177 passengers per OD pair, corresponding to an increase of approximately 17.8\%. The OD standard deviation increases from 1.3790 to 1.5926, while the proportion of zero OD flows decreases from 70.25\% to 67.80\%. Mean station outflow also increases, from 7.3118 to 8.6127 passengers.

\begin{table}[H]
\centering
\caption{Passenger-demand characteristics in undisrupted and network-level disrupted intervals within the modelling population. An interval is classified as disrupted when the network-wide mean station-call delay exceeds 60 seconds. Only intervals with valid station-call observations are included.}
\resizebox{\textwidth}{!}{
\begin{tabular}{lrrrrr}
\toprule
\textbf{Condition} &
\textbf{Intervals} &
\textbf{Mean OD demand} &
\textbf{OD standard deviation} &
\textbf{Zero-flow rate (\%)} &
\textbf{Mean station outflow} \\
\midrule
Undisrupted & 13,044 & 0.6093 & 1.3790 & 70.25 & 7.3118 \\
Disrupted   & 1,241  & 0.7177 & 1.5926 & 67.80 & 8.6127 \\
\bottomrule
\end{tabular}}
\label{tab:disruption_demand_shift}
\end{table}

This comparison indicates that network-level disrupted intervals are associated with a shifted and more variable demand regime within the same population used for forecasting. However, the result is descriptive rather than causal. Passenger demand and service reliability both vary systematically with operating conditions and time of day, and periods of high passenger demand may themselves contribute to service delays. The observed differences should therefore not be interpreted as evidence that delays alone cause the increase in demand. The following analysis examines whether the larger reconciliation gains observed under disruption are associated with greater base-forecast coherence violations between the station and OD levels.
 
\subsubsection{Base-Forecast Coherence Violation and Reconciliation Gains}
\label{sec:disruption_disagreement}

Following the definition in Section~\ref{sec:evaluation_metrics}, we analyse the interval-level coherence violation of the unreconciled base forecasts. Specifically, we use $D_t \equiv D^{(\mathrm{base})}_{t,1}$, corresponding to the one-step discrepancy between the direct station-level base forecast, covering both outflow and inflow, and the station outflow and inflow implied by the independently generated OD base forecast. A value of zero indicates that the two forecasts satisfy both aggregation constraints exactly. When interpreting its variation across intervals and operating conditions, we refer to this quantity as cross-level forecast disagreement. Unlike a forecasting-error metric, $D_t$ does not involve observed passenger demand. Averaged over the complete test set, $D_t$ equals the base coherence-violation MSE of 0.6501 reported in Table~\ref{tab:single_step_hierarchy_performance}.

The interval-level forecasting gain obtained through FCR is defined as
\begin{equation}
G_t =
\frac{1}{N_{\mathrm{od}}}
\left\|
\mathbf{y}_{t+1}
-
\widehat{\mathbf{y}}_{t+1}
\right\|_2^2
-
\frac{1}{N_{\mathrm{od}}}
\left\|
\mathbf{y}_{t+1}
-
\widetilde{\mathbf{y}}_{t+1}
\right\|_2^2,
\label{eq:interval_reconciliation_gain}
\end{equation}
where a positive value indicates that reconciliation reduces OD forecasting error relative to the unreconciled base forecast.

For this diagnostic, disruption conditions are defined at the interval level, and the forecasting metrics are calculated over the complete station and OD vectors in every selected interval. The values in Table~\ref{tab:cross_level_disagreement} therefore differ from those in Tables~\ref{tab:cancellation_disruptions} and \ref{tab:representative_delay_disruptions}, which condition and evaluate individual OD observations. An interval is classified as undisrupted when no cancellation is recorded and neither origin- nor destination-side delay exceeds 60 seconds. Small positive delays below this threshold may therefore remain in the undisrupted subset.

\begin{table}[H]
\centering
\caption{Base-forecast accuracy, mean base coherence violation computed jointly
over station outflow and inflow, and FCR forecasting gain under representative
operational conditions. The Spearman coefficient measures the interval-level
association between base coherence violation and the absolute reduction in OD
forecasting error.}
\resizebox{\textwidth}{!}{%
\begin{tabular}{lccccccc}
\toprule
\textbf{Condition} &
\textbf{Intervals} &
\textbf{Base station MSE} &
\textbf{Base OD MSE} &
\textbf{\makecell{Mean coherence\\violation}} &
\textbf{FCR OD MSE} &
\textbf{Absolute gain} &
\(\boldsymbol{\rho}\) \\
\midrule
Undisrupted
    & 1,690 & 8.6398 & 0.7027 & 0.5377 & 0.6901 & 0.0125 & 0.218 \\
Origin delay $>60$ s
    & 801 & 11.1066 & 0.9089 & 0.8579 & 0.8868 & 0.0221 & 0.307 \\
Destination delay $>60$ s
    & 769 & 10.6956 & 0.8826 & 0.8002 & 0.8612 & 0.0214 & 0.303 \\
Origin cancellations $>0$
    & 452 & 11.5238 & 0.9721 & 0.9367 & 0.9456 & 0.0265 & 0.342 \\
Destination cancellations $>0$
    & 454 & 11.4961 & 0.9749 & 0.9342 & 0.9491 & 0.0258 & 0.337 \\
\bottomrule
\end{tabular}%
}
\label{tab:cross_level_disagreement}
\end{table}

Relative to undisrupted intervals, the mean base coherence violation is 59.5\% higher during origin-delay intervals exceeding 60 seconds and 48.8\% higher during destination-delay intervals. Under origin- and destination-side cancellations, the mean violation is approximately 74\% higher than in undisrupted intervals. The absolute reduction in OD forecasting error achieved through FCR is approximately 1.7 times larger under the representative delay conditions and about twice as large under cancellation conditions as in the undisrupted subset.

The interval-level base coherence violation is also positively associated with the forecasting gain obtained through reconciliation. Spearman correlations range from 0.303 to 0.342 across the representative disruption subsets and are statistically significant at $p<0.001$. These moderate associations indicate that greater structural disagreement between the independently generated station- and OD-level forecasts is related to larger reconciliation gains, while also showing that coherence violation is not the sole determinant of forecasting improvement.

To examine this relationship independently of predefined disruption categories, we divide all test intervals into terciles according to their base-forecast coherence violation. Table~\ref{tab:disagreement_quantiles} shows that the mean absolute reconciliation gain increases consistently across the three groups. Intervals in the highest-violation tercile obtain a mean OD-MSE reduction of 0.0371, compared with 0.0019 in the lowest-violation tercile.

\begin{table}[H]
\centering
\caption{Reconciliation performance across terciles of the combined
outflow--inflow base-forecast coherence violation in the complete test set.}
\resizebox{\textwidth}{!}{%
\begin{tabular}{lccccc}
\toprule
\textbf{\makecell{Coherence-violation \\group}} &
\textbf{Intervals} &
\textbf{\makecell{Mean coherence\\violation}} &
\textbf{Base OD MSE} &
\textbf{FCR OD MSE} &
\textbf{Absolute gain} \\
\midrule
Low    & 951 & 0.1345 & 0.3924 & 0.3905 & 0.0019 \\
Medium & 950 & 0.3567 & 0.7575 & 0.7484 & 0.0091 \\
High   & 951 & 1.4588 & 1.1970 & 1.1599 & 0.0371 \\
\bottomrule
\end{tabular}%
}
\label{tab:disagreement_quantiles}
\end{table}

Together, these results provide an empirical interpretation of the larger reconciliation gains observed under disruption. The representative disruption conditions are associated with a shifted demand distribution and larger coherence violations between the independently generated station- and OD-level base forecasts. Across both the disruption-defined subsets and the complete test set, larger base coherence violations are associated with greater reductions in OD forecasting error after reconciliation. This suggests that reconciliation is particularly useful when the two forecasting levels provide divergent but complementary information.

The observed association does not imply that coherence violation alone causes the forecasting improvement. FCR neither models the disruption process explicitly nor adapts its parameters during inference. Rather, the fixed reconciler uses the information contained in both forecasting levels to produce more accurate and structurally coherent predictions.

\begin{figure}[H]
    \centering
    \includegraphics[width=\textwidth]
    {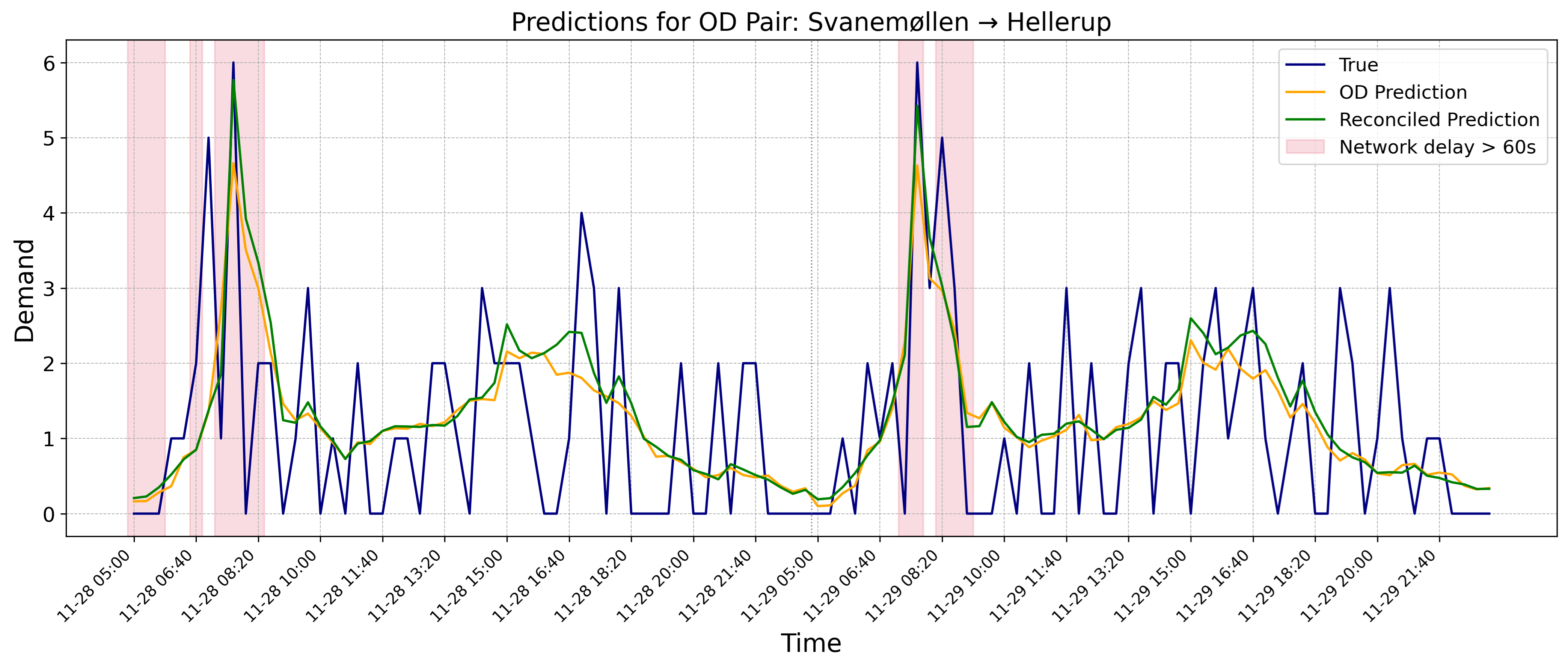}
    \caption{Observed and one-step-ahead predicted demand for the Svanemøllen--Hellerup OD pair over a representative contiguous test window. Shaded regions indicate 20-minute intervals in which the network-wide mean delay across valid station calls exceeds 60 seconds. The figure compares observed demand with the unreconciled base OD forecast and the FCR-reconciled forecast. The window and OD pair were selected independently of forecasting error and reconciliation gain.}
    \label{fig:disruption_prediction}
\end{figure}

Figure~\ref{fig:disruption_prediction} complements the aggregate analysis with an observed-versus-predicted comparison for the Svanemøllen--Hellerup OD pair over a contiguous two-day test window. The shaded regions indicate 20-minute intervals in which the network-wide mean delay across valid station calls exceeds 60 seconds. This network-level criterion identifies periods of broadly degraded operating conditions and differs from the origin- and destination-specific masks used in Tables~\ref{tab:cancellation_disruptions} and \ref{tab:representative_delay_disruptions}.

The operational records do not provide consistent incident identifiers, and multiple delayed services may overlap within the same 20-minute interval. Consequently, the figure is presented as a representative disrupted time window rather than as a reconstruction of a single isolated incident.

As shown in Figure~\ref{fig:disruption_prediction}, both the base and reconciled forecasts reproduce the broader temporal demand profile, including the principal demand peaks occurring during the network-level disruption periods. The observed OD series remains more volatile and sparse than either forecast at the 20-minute resolution. FCR produces moderate adjustments to the base OD forecast, consistent with its role as a cross-level correction mechanism rather than a disruption-specific forecasting model.
 
\subsubsection{Multi-Step Reconciliation under Disruption}

The multi-step evaluation contains 2,847 valid forecast origins, five fewer than
the one-step evaluation because the final five origins of the test period do not
have complete target sequences through \(T+6\). The same valid multi-step
population is used across all six forecast horizons. Tables~\ref{tab:multistep_cancellation}
and~\ref{tab:multistep_representative_delay} report the percentage reduction in
OD MSE obtained by S-FCR under the same representative operational conditions.
The ``OD observations'' column reports the number of OD-pair forecasts satisfying
each condition within this common multi-step evaluation population.

\begin{table}[H]
\centering
\caption{S-FCR reduction in OD MSE (\%) across forecast horizons under train
cancellation conditions.}
\resizebox{\textwidth}{!}{
\begin{tabular}{lccccccc}
\toprule
\textbf{Condition} &
\textbf{T+1} &
\textbf{T+2} &
\textbf{T+3} &
\textbf{T+4} &
\textbf{T+5} &
\textbf{T+6} &
\textbf{OD observations} \\
\midrule
Origin cancellations $=0$
    & 4.06 & 3.99 & 3.84 & 3.66 & 3.67 & 4.00 & 372,484 \\
Origin cancellations $>0$
    & 4.82 & 6.65 & 4.10 & 3.42 & 2.61 & 2.72 & 3,320 \\
\midrule
Destination cancellations $=0$
    & 4.02 & 4.00 & 3.85 & 3.67 & 3.65 & 3.98 & 372,625 \\
Destination cancellations $>0$
    & 8.09 & 5.75 & 2.63 & 2.34 & 5.56 & 6.56 & 3,179 \\
\bottomrule
\end{tabular}}
\label{tab:multistep_cancellation}
\end{table}

Under cancellation-free conditions, S-FCR reduces OD MSE by approximately
3.7\%--4.1\% across the six horizons. Reconciliation also improves the forecasts
at every horizon when cancellations are recorded. For origin-side cancellations,
the reductions range from 2.61\% to 6.65\%, while under destination-side
cancellations they range from 2.34\% to 8.09\%. The variation across horizons is
non-monotonic and does not indicate a systematic increase or decrease in
reconciliation benefit with forecast horizon.

\begin{table}[H]
\centering
\caption{S-FCR reduction in OD MSE (\%) across forecast horizons under
representative delay conditions.}
\resizebox{\textwidth}{!}{
\begin{tabular}{lccccccc}
\toprule
\textbf{Condition} &
\textbf{T+1} &
\textbf{T+2} &
\textbf{T+3} &
\textbf{T+4} &
\textbf{T+5} &
\textbf{T+6} &
\textbf{OD observations} \\
\midrule
Origin delay $=0$ s
    & 4.02 & 4.02 & 3.84 & 3.70 & 3.70 & 4.05 & 339,550 \\
Origin delay $>60$ s
    & 6.06 & 5.14 & 4.18 & 4.68 & 2.78 & 4.31 & 3,208 \\
\midrule
Destination delay $=0$ s
    & 3.88 & 3.88 & 3.75 & 3.68 & 3.74 & 3.95 & 339,937 \\
Destination delay $>60$ s
    & 9.86 & 5.99 & 4.49 & 2.50 & 2.82 & 3.10 & 3,074 \\
\bottomrule
\end{tabular}}
\label{tab:multistep_representative_delay}
\end{table}

For origin-side delays exceeding 60 seconds, S-FCR reduces OD MSE by
2.78\%--6.06\% across the six horizons. For destination-side delays, the
reductions range from 2.50\% to 9.86\%. Reconciliation therefore improves upon
the unreconciled forecasts at every examined horizon, although the magnitude of
the improvement varies across the prediction window.

Overall, the multi-step results show that hierarchical reconciliation remains
beneficial under representative train-delay and cancellation conditions.
Together with the preceding coherence-violation analysis, the results indicate
that larger reconciliation gains tend to occur when the independently generated
station- and OD-level base forecasts exhibit greater structural disagreement.
Results for the less frequent 180- and 300-second delay thresholds are reported
in Appendix~\ref{app:additional_disruption_results} as exploratory sensitivity
analyses and are not used to infer a monotonic relationship between disruption
severity and reconciliation performance.

\section{Conclusion}

This study introduced a hierarchical forecast reconciliation framework for urban rail transit demand that jointly reconciles station-level and origin--destination flow predictions. The framework combines SDT-GRU for station-level forecasting, mGraphSAGE for OD-level forecasting, and a neural Fully Connected Reconciler (FCR) that learns a nonlinear mapping from incoherent base forecasts to coherent hierarchical predictions. Exact structural coherence is guaranteed by construction through the fixed summing matrix $\mathbf{S}$.

Across the one-step and multi-step settings, reconciliation improved OD-level accuracy relative to the unreconciled base forecasts while eliminating cross-level coherence violations up to numerical precision. Under standard one-step conditions, FCR performed competitively with the evaluated MinT-based methods, while MinT-Sample achieved the best overall accuracy on several metrics. Under representative disruption conditions involving train cancellations and
delays exceeding 60 seconds, reconciliation continued to improve OD forecasting
accuracy, with FCR yielding larger point reductions than MinT-Sample in the
affected subsets. In the multi-step evaluation, S-FCR achieved lower OD, station, and full-hierarchy MSE than MinT-Sample at every horizon from T+1 to T+6, without requiring separately learned parameters for each horizon and even though MinT-Sample used horizon-specific covariance estimates. An oracle analysis in which observed station demand replaced the station-level input to the trained FCR reduced OD MSE by approximately 34\%, quantifying the reconciliation potential that remains as station-level forecasting accuracy improves.

A joint multi-task baseline, predicting both levels within a single model that couples them during encoding, was additionally evaluated on the same data under a matched parameter budget. Without a consistency term, its coherence violation exceeded that of the independently generated base forecasts, indicating that a shared representation does not by itself induce structural agreement between the two prediction heads. A soft aggregation-consistency penalty reduced the violation by approximately 30\%, but it remained many orders of magnitude above the near-exact coherence obtained by reconstructing forecasts through $\mathbf{S}$. The baseline also did not improve forecasting accuracy, although that comparison rests on a single untuned architecture and warrants more caution than the coherence result.

The diagnostic analyses indicate when reconciliation is most beneficial. Network-level disrupted intervals exhibited higher and more variable passenger demand, and the representative delay and cancellation subsets showed larger coherence violations between the station-level base forecasts and the station demand implied by the OD base forecasts. At the interval level, these violations were positively, although moderately, associated with the reduction in OD forecasting error obtained through reconciliation. Reconciliation therefore appears most useful when the two base models provide divergent but complementary information, serving both as a coherence-enforcement procedure and as a cross-level correction mechanism. These results are associative rather than causal and do not imply that FCR explicitly models the disruption process or adapts its parameters during inference.

From a practical standpoint, the framework provides a post-processing solution that can be integrated with existing forecasting pipelines without retraining the underlying models, and its modular formulation permits different station- and OD-level architectures to be combined, although broader validation is required before operational deployment.

Several limitations point to directions for future work. The case study is limited to a single 12-station subnetwork of the Copenhagen S-train system, comprising 132 directed OD pairs. Although this setting supports a detailed evaluation under operational disruptions, it establishes neither cross-city generalisability nor computational scalability to substantially larger rail networks, and comparable integrated data on demand and station-call records were not available from additional cities. Extending the full pipeline to the complete Copenhagen network would assess within-city scalability, while evaluation on other URT systems would be required to assess transfer across network topologies and operating environments. The disruption observations also follow a naturally imbalanced distribution, and the less frequent and more severe conditions have limited empirical support; evaluation over longer observation periods or controlled disruption scenarios would enable a more robust assessment under rare operating conditions.

Further directions follow from the framework itself. It currently produces deterministic point forecasts, limiting its use in risk-aware planning; extending reconciliation to forecast distributions rather than point estimates is therefore an important next step. The joint-baseline comparison reported here is an initial assessment of joint learning against post-processing reconciliation, and a broader comparison encompassing tuned architectures, multi-step and disruption settings, and repeated training runs would give a more complete account of that trade-off. Additional directions include online reconciliation that adapts to evolving operating regimes without full retraining, structurally constrained reconcilers that reduce parameter growth from $O(N^4)$ to $O(N^2)$, and graph-aware or sequence-aware architectures that model spatial, temporal, and hierarchical dependencies jointly rather than only through post-processing. Taken together, these extensions would advance the framework toward a more scalable, uncertainty-aware, and operationally validated forecasting system for urban rail transit.

\appendix

\section{Temporal Distribution of Operational Disruptions}
\label{app:disruption_distribution}

\begin{figure}[htbp]
    \centering
    \includegraphics[width=\textwidth]{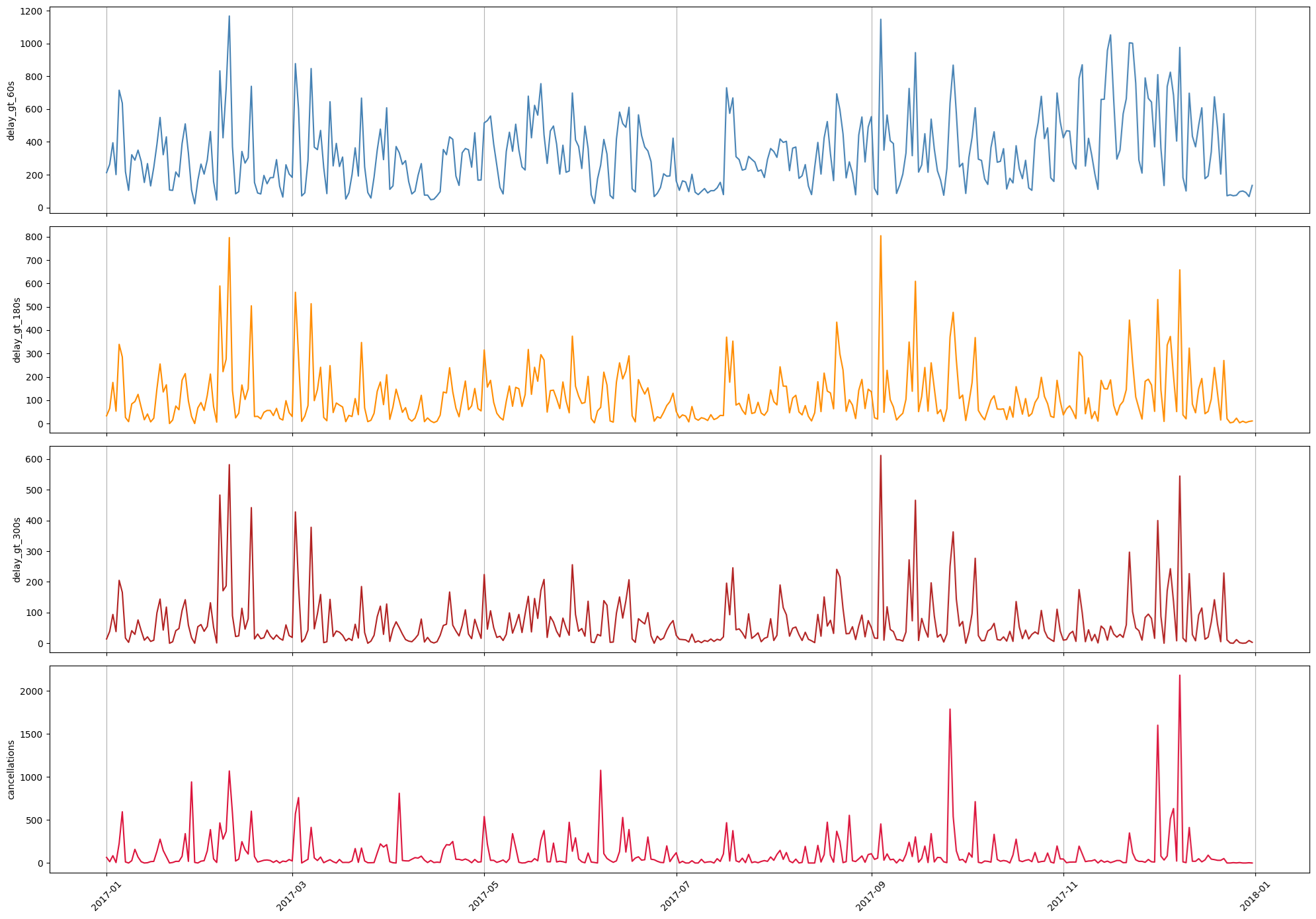}
    \caption{Daily frequency of operational disruption observations across
    the 12-station study network in 2017. Counts represent individual
    station-call records. The delay thresholds are cumulative: station calls
    with delays exceeding 300 seconds are also included in the counts for
    delays exceeding 180 and 60 seconds. Cancellation counts are reported
    separately.}
    \label{fig:daily_disruption}
\end{figure}

Figure~\ref{fig:daily_disruption} shows the daily frequency of
delay and cancellation observations across the 12-station study network
during 2017. The figure complements Table~\ref{tab:disruption_frequency} by
showing how disruption observations are distributed over time rather than
only reporting their aggregate frequency.

\section{Additional Disruption Sensitivity Analyses}
\label{app:additional_disruption_results}

This appendix reports supplementary results for less frequently observed
severe-delay conditions and for non-operational atypical conditions,
including holidays and adverse weather. These analyses are retained for
completeness but are treated as exploratory because some subsets contain
substantially fewer observations than the representative disruption
conditions analysed in Section~\ref{sec:disruption_results}.

In the following tables, ``OD observations'' denotes the number of OD-pair
forecasts included in each error calculation. It should not be interpreted
as the number of independent disruption events, because multiple OD
observations may correspond to the same 20-minute forecasting interval.
Station-call-level disruption frequencies are reported in
Table~\ref{tab:disruption_frequency}.

\subsection{One-Step Reconciliation under Severe Delay Conditions}
\label{app:severe_delay_one_step}

Tables~\ref{tab:appendix_origin_delay} and
\ref{tab:appendix_destination_delay} report one-step results for mean delays
exceeding 180 and 300 seconds. FCR improves upon the unreconciled OD forecasts
in all four subsets and achieves larger reductions than MinT-Sample.
However, these results are not used to establish a monotonic relationship
between disruption severity and reconciliation performance because the
qualifying subsets are relatively small.

\begin{table}[H]
\centering
\caption{Exploratory one-step OD forecasting performance under severe
origin-delay conditions. Reductions are measured relative to the
unreconciled base OD forecast.}
\resizebox{\textwidth}{!}{
\begin{tabular}{lcccccc}
\toprule
\textbf{Condition} &
\textbf{Base OD} &
\textbf{FCR} &
\textbf{MinT} &
\textbf{FCR reduction (\%)} &
\textbf{MinT reduction (\%)} &
\textbf{OD observations} \\
\midrule
Origin delay $>180$ s
    & 0.8783 & 0.8196 & 0.8647 & 6.68 & 1.55 & 727 \\
Origin delay $>300$ s
    & 0.9879 & 0.9316 & 0.9524 & 5.70 & 3.59 & 290 \\
\bottomrule
\end{tabular}}
\label{tab:appendix_origin_delay}
\end{table}

\begin{table}[H]
\centering
\caption{Exploratory one-step OD forecasting performance under severe
destination-delay conditions. Reductions are measured relative to the
unreconciled base OD forecast.}
\resizebox{\textwidth}{!}{
\begin{tabular}{lcccccc}
\toprule
\textbf{Condition} &
\textbf{Base OD} &
\textbf{FCR} &
\textbf{MinT} &
\textbf{FCR reduction (\%)} &
\textbf{MinT reduction (\%)} &
\textbf{OD observations} \\
\midrule
Destination delay $>180$ s
    & 1.0679 & 0.9988 & 1.0393 & 6.47 & 2.67 & 662 \\
Destination delay $>300$ s
    & 0.9145 & 0.8402 & 0.8676 & 8.12 & 5.12 & 291 \\
\bottomrule
\end{tabular}}
\label{tab:appendix_destination_delay}
\end{table}

The severe-delay subsets suggest that reconciliation can remain beneficial
under highly atypical operating conditions. Nevertheless, the differences
between the 180- and 300-second thresholds are not monotonic, and the small
number of qualifying OD observations limits the strength of conclusions that
can be drawn from these comparisons.

\subsection{Multi-Step Reconciliation under Severe Delay Conditions}
\label{app:severe_delay_multi_step}

Tables~\ref{tab:appendix_multistep_origin_delay} and
\ref{tab:appendix_multistep_destination_delay} report S-FCR reductions in OD
MSE across the six forecast horizons for the severe-delay subsets.

\begin{table}[H]
\centering
\caption{Exploratory S-FCR reduction in OD MSE (\%) across forecast horizons
under severe origin-delay conditions.}
\resizebox{\textwidth}{!}{
\begin{tabular}{lccccccc}
\toprule
\textbf{Condition} &
\textbf{T+1} &
\textbf{T+2} &
\textbf{T+3} &
\textbf{T+4} &
\textbf{T+5} &
\textbf{T+6} &
\textbf{OD observations} \\
\midrule
Origin delay $>180$ s
    & 3.97 & 2.41 & 2.52 & 6.43 & 4.89 & 6.25 & 727 \\
Origin delay $>300$ s
    & 6.31 & 3.72 & 8.06 & 11.11 & 7.35 & 9.74 & 290 \\
\bottomrule
\end{tabular}}
\label{tab:appendix_multistep_origin_delay}
\end{table}

\begin{table}[H]
\centering
\caption{Exploratory S-FCR reduction in OD MSE (\%) across forecast horizons
under severe destination-delay conditions.}
\resizebox{\textwidth}{!}{
\begin{tabular}{lccccccc}
\toprule
\textbf{Condition} &
\textbf{T+1} &
\textbf{T+2} &
\textbf{T+3} &
\textbf{T+4} &
\textbf{T+5} &
\textbf{T+6} &
\textbf{OD observations} \\
\midrule
Destination delay $>180$ s
    & 12.70 & 8.37 & 5.22 & 3.08 & 4.56 & 4.84 & 662 \\
Destination delay $>300$ s
    & 17.45 & 4.22 & 7.38 & 6.57 & 9.32 & 8.44 & 291 \\
\bottomrule
\end{tabular}}
\label{tab:appendix_multistep_destination_delay}
\end{table}

S-FCR improves upon the unreconciled forecasts at every horizon in these
subsets. The magnitude of the reduction varies considerably across horizons,
particularly for destination delays exceeding 300 seconds. Given the limited
support of these conditions, the large individual reductions, including the
17.45\% value at T+1, should be interpreted as exploratory rather than as
stable estimates of performance under severe disruptions.

\subsection{Holiday-Period Results}
\label{app:holiday_results}

Table~\ref{tab:appendix_holiday_results} reports one-step FCR performance
during four holiday periods. Reconciliation improves OD forecasting accuracy
in all four subsets, although the magnitude of the reduction varies across
holiday types.

\begin{table}[H]
\centering
\caption{Exploratory one-step OD forecasting performance during holiday
periods.}
\begin{tabular}{lcccc}
\toprule
\textbf{Holiday period} &
\textbf{Base OD} &
\textbf{FCR OD} &
\textbf{Reduction (\%)} &
\textbf{OD observations} \\
\midrule
Easter break & 0.4759 & 0.4611 & 3.10 & 14,520 \\
Winter break & 0.6530 & 0.6402 & 1.96 & 7,260 \\
Summer break & 0.6335 & 0.6174 & 2.54 & 72,600 \\
Autumn break & 0.6000 & 0.5952 & 0.79 & 7,260 \\
\bottomrule
\end{tabular}
\label{tab:appendix_holiday_results}
\end{table}

These results indicate that reconciliation remains beneficial during
seasonally atypical periods. Because the holiday categories differ in
duration, travel purpose, and temporal coverage, the differences between
their reported reductions should not be interpreted as causal effects of
specific holiday types.

\subsection{Weather-Related Results}
\label{app:weather_results}

Tables~\ref{tab:appendix_precipitation_results} and
\ref{tab:appendix_temperature_wind_results} report one-step FCR performance
under precipitation, temperature, and wind conditions.

\begin{table}[H]
\centering
\caption{Exploratory one-step OD forecasting performance under precipitation
conditions.}
\begin{tabular}{lcccc}
\toprule
\textbf{Condition} &
\textbf{Base OD} &
\textbf{FCR OD} &
\textbf{Reduction (\%)} &
\textbf{OD observations} \\
\midrule
No rain & 0.7655 & 0.7497 & 2.07 & 290,136 \\
Rainy weather & 0.8389 & 0.8220 & 2.01 & 86,328 \\
Heavy rain ($>3$ mm) & 0.7470 & 0.7109 & 4.83 & 792 \\
\midrule
No snow & 0.7801 & 0.7643 & 2.02 & 369,996 \\
Snowfall $> 1$ cm & 0.5849 & 0.5469 & 6.51 & 792 \\
\bottomrule
\end{tabular}
\label{tab:appendix_precipitation_results}
\end{table}

\begin{table}[H]
\centering
\caption{Exploratory one-step OD forecasting performance under temperature
and wind conditions.}
\begin{tabular}{lcccc}
\toprule
\textbf{Condition} &
\textbf{Base OD} &
\textbf{FCR OD} &
\textbf{Reduction (\%)} &
\textbf{OD observations} \\
\midrule
Temperature $<0^\circ$C
    & 0.6152 & 0.6004 & 2.41 & 9,768 \\
Temperature $>20^\circ$C
    & 1.0205 & 1.0128 & 0.75 & 11,088 \\
\midrule
Wind speed $<10$ km/h
    & 0.6824 & 0.6676 & 2.17 & 54,120 \\
Wind speed $>15$ km/h
    & 0.8006 & 0.7844 & 2.02 & 237,336 \\
\bottomrule
\end{tabular}
\label{tab:appendix_temperature_wind_results}
\end{table}

Reconciliation improves OD accuracy in all reported weather subsets. The
larger reductions observed for heavy rain and snow are based on only 792 OD
observations each and should therefore be interpreted cautiously. These
results are retained as supplementary evidence that reconciliation does not
systematically degrade forecasts under atypical weather, rather than as
evidence that its benefits increase with weather severity.










\bibliographystyle{elsarticle-num-names} 
\bibliography{cas-refs}

@article{Li2024Long-TermInformer,
    title = {{Long-Term Passenger Flow Forecasting for Rail Transit Based on Complex Networks and Informer}},
    year = {2024},
    journal = {Sensors},
    author = {Li, Dekui and Du, Shubo and Hou, Yuru},
    number = {21},
    month = {11},
    volume = {24},
    publisher = {Multidisciplinary Digital Publishing Institute (MDPI)},
    doi = {10.3390/s24216894},
    issn = {14248220}
}

@article{Lu2025AFlow,
  title={A novel integrative prediction framework for metro passenger flow},
  author={Lu, Wenbo and Zhang, Yong and Vu, Hai L and Xu, Jinhua and Li, Peikun},
  journal={Journal of Intelligent Transportation Systems},
  pages={1--26},
  year={2025},
  publisher={Taylor \& Francis}
}

@article{Lv2024AnMechanism,
    title = {{An origin–destination passenger flow prediction system based on convolutional neural network and passenger source-based attention mechanism}},
    year = {2024},
    journal = {Expert Systems with Applications},
    author = {Lv, Sirui and Wang, Kaipeng and Yang, Hu and Wang, Pu},
    month = {3},
    pages = {121989},
    volume = {238},
    publisher = {Pergamon},
    doi = {10.1016/J.ESWA.2023.121989},
    issn = {0957-4174}
}

@article{Yang2024AreForecasting,
    title = {{Are Graphs and GCNs necessary for short-term metro ridership forecasting?}},
    year = {2024},
    journal = {Expert Systems with Applications},
    author = {Yang, Qiong and Xu, Xianghua and Wang, Zihang and Yu, Juan and Hu, Xiaodong},
    month = {11},
    volume = {254},
    publisher = {Elsevier Ltd},
    doi = {10.1016/j.eswa.2024.124431},
    issn = {09574174}
}

@article{Halyal2022ForecastingData,
    title = {{Forecasting public transit passenger demand: With neural networks using APC data}},
    year = {2022},
    journal = {Case Studies on Transport Policy},
    author = {Halyal, Shivaraj and Mulangi, Raviraj H. and Harsha, M. M.},
    number = {2},
    month = {6},
    pages = {965--975},
    volume = {10},
    publisher = {Elsevier},
    doi = {10.1016/J.CSTP.2022.03.011},
    issn = {2213-624X}
}

@article{Lu2023MOHP-EC:Flow,
    title = {{MOHP-EC: A Multiobjective Hierarchical Prediction Framework for Urban Rail Transit Passenger Flow}},
    year = {2023},
    journal = {IEEE Intelligent Transportation Systems Magazine},
    author = {Lu, Wenbo and Xu, Jinhua and Zhang, Yong and Wang, Ting and Li, Peikun},
    number = {4},
    month = {7},
    pages = {86--105},
    volume = {15},
    publisher = {Institute of Electrical and Electronics Engineers Inc.},
    doi = {10.1109/MITS.2023.3242465},
    issn = {19411197}
}

@article{Hyndman2011OptimalSeries,
  author    = {Hyndman, Rob J. and Ahmed, Roman A. and Athanasopoulos, George and Shang, Han Lin},
  title     = {Optimal combination forecasts for hierarchical time series},
  journal   = {Computational Statistics \& Data Analysis},
  volume    = {55},
  number    = {9},
  pages     = {2579--2589},
  year      = {2011},
  doi       = {10.1016/j.csda.2011.03.006}
}

@article{Athanasopoulos2024ForecastReview,
  author    = {Athanasopoulos, George and Hyndman, Rob J. and Kourentzes, Nikolaos and Panagiotelis, Anastasios},
  title     = {Forecast reconciliation: A review},
  journal   = {International Journal of Forecasting},
  volume    = {40},
  number    = {2},
  pages     = {430--456},
  year      = {2024},
  doi       = {10.1016/j.ijforecast.2023.10.010}
}

@article{Hollyman2021UnderstandingReconciliation,
  author    = {Hollyman, Ross and Petropoulos, Fotios and Tipping, Michael E.},
  title     = {Understanding forecast reconciliation},
  journal   = {European Journal of Operational Research},
  volume    = {294},
  number    = {1},
  pages     = {149--160},
  year      = {2021},
  doi       = {10.1016/j.ejor.2021.01.017}
}

@article{Wickramasuriya2019OptimalReconciliation,
  author    = {Wickramasuriya, Shanika L. and Athanasopoulos, George and Hyndman, Rob J.},
  title     = {Optimal forecast reconciliation for hierarchical and grouped time series through trace minimization},
  journal   = {Journal of the American Statistical Association},
  volume    = {114},
  number    = {526},
  pages     = {804--819},
  year      = {2019},
  doi       = {10.1080/01621459.2018.1448825}
}

@article{Hyndman2016WLS,
  author    = {Hyndman, Rob J. and Lee, Alan J. and Wang, Earo},
  title     = {Fast computation of reconciled forecasts for hierarchical and grouped time series},
  journal   = {Computational Statistics \& Data Analysis},
  volume    = {97},
  pages     = {16--32},
  year      = {2016},
  doi       = {10.1016/j.csda.2015.11.007}
}

@article{Spiliotis2021HierarchicalML,
  author    = {Spiliotis, Evangelos and Abolghasemi, Mahdi and Hyndman, Rob J. and Petropoulos, Fotios and Assimakopoulos, Vassilios},
  title     = {Hierarchical forecast reconciliation with machine learning},
  journal   = {Applied Soft Computing},
  volume    = {112},
  pages     = {107756},
  year      = {2021},
  doi       = {10.1016/j.asoc.2021.107756}
}

@misc{Burba2021Encoder,
  author    = {Burba, Danny and Chen, Tianqi},
  title     = {A trainable reconciliation method for hierarchical time-series},
  year      = {2021},
  note      = {arXiv preprint arXiv:2101.01329},
  url       = {https://arxiv.org/abs/2101.01329}
}

@inproceedings{Rangapuram2021EndToEnd,
  author    = {Rangapuram, Syama Sundar and Werner, Luciana Deep and Benidis, Konstantinos and Mercado, Pedro and Gasthaus, Jan and Januschowski, Tim},
  title     = {End-to-end learning of coherent probabilistic forecasts for hierarchical time series},
  booktitle = {Proceedings of the 38th International Conference on Machine Learning},
  pages     = {8832--8843},
  year      = {2021},
  url       = {http://proceedings.mlr.press/v139/rangapuram21a.html}
}

@article{Wang2021HypergraphMetro,
  author  = {Wang, Junbo and Zhang, Yong and Wei, Yi and Hu, Yihang and Piao, Xingchen and Yin, Baocai},
  title   = {Metro passenger flow prediction via dynamic hypergraph convolution networks},
  journal = {IEEE Transactions on Intelligent Transportation Systems},
  volume  = {22},
  number  = {12},
  pages   = {7891--7903},
  year    = {2021}
}

@article{Liu2020PhysicalVirtual,
  author  = {Liu, Lingbo and Chen, Jingxing and Wu, Hefeng and Zhen, Jiajie and Li, Guanbin and Lin, Liang},
  title   = {Physical-virtual collaboration modeling for intra- and inter-station metro ridership prediction},
  journal = {IEEE Transactions on Intelligent Transportation Systems},
  volume  = {23},
  number  = {4},
  pages   = {3377--3391},
  year    = {2020}
}

@article{Ma2019ParallelBiLSTM,
  author  = {Ma, Xueyao and Zhang, Jian and Du, Bo and Ding, Chuan and Sun, Li},
  title   = {Parallel architecture of convolutional bi-directional {LSTM} neural networks for network-wide metro ridership prediction},
  journal = {IEEE Transactions on Intelligent Transportation Systems},
  volume  = {20},
  number  = {6},
  pages   = {2278--2288},
  year    = {2019}
}

@article{Bao2022AttentionMultiview,
  author  = {Bao, Jie and Kang, Jian and Yang, Zhiyong and Chen, Xiqun},
  title   = {Forecasting network-wide multi-step metro ridership with an attention-weighted multi-view graph to sequence learning approach},
  journal = {Expert Systems with Applications},
  volume  = {210},
  pages   = {118475},
  year    = {2022}
}

@article{Fang2024DualView,
  author  = {Fang, Hao and Chen, Ching-Hung and Hwang, Fong-Jou and Chang, Chia-Chen and Chang, Chun-Cheng},
  title   = {Metro station functional clustering and dual-view recurrent graph convolutional network for metro passenger flow prediction},
  journal = {Expert Systems with Applications},
  volume  = {247},
  pages   = {122550},
  year    = {2024}
}

@article{hornik1991approximation,
  title={Approximation capabilities of multilayer feedforward networks},
  author={Hornik, Kurt},
  journal={Neural networks},
  volume={4},
  number={2},
  pages={251--257},
  year={1991},
  publisher={Elsevier}
}

@article{nguyen2026multi,
  title={Multi-Graph Inductive Representation Learning for large-scale Urban Rail demand prediction under disruptions},
  author={Nguyen, Dang Viet Anh and Flensburg, J Victor and Cerreto, Fabrizio and Pascariu, Bianca and Pellegrini, Paola and Azevedo, Carlos Lima and Rodrigues, Filipe},
  journal={Computers \& Industrial Engineering},
  pages={111924},
  year={2026},
  publisher={Elsevier}
}

@article{huang2024mtlmetro,
  title={MTLMetro: A deep multi-task learning model for metro passenger demands prediction},
  author={Huang, Hao and Mao, Jiannan and Liu, Ronghui and Lu, Weike and Tang, Tianli and Liu, Lan},
  journal={IEEE Transactions on Intelligent Transportation Systems},
  volume={25},
  number={9},
  pages={11805--11820},
  year={2024},
  publisher={IEEE}
}

@article{liu2022online,
  title={Online metro origin-destination prediction via heterogeneous information aggregation},
  author={Liu, Lingbo and Zhu, Yuying and Li, Guanbin and Wu, Ziyi and Bai, Lei and Lin, Liang},
  journal={IEEE Transactions on Pattern Analysis and Machine Intelligence},
  volume={45},
  number={3},
  pages={3574--3589},
  year={2022},
  publisher={IEEE}
}

@article{xu2023adaptive,
  title={Adaptive feature fusion networks for origin-destination passenger flow prediction in metro systems},
  author={Xu, Yuhang and Lyu, Yan and Xiong, Guangwei and Wang, Shuyu and Wu, Weiwei and Cui, Helei and Luo, Junzhou},
  journal={IEEE Transactions on Intelligent Transportation Systems},
  volume={24},
  number={5},
  pages={5296--5312},
  year={2023},
  publisher={IEEE}
}

@article{zhao2025multi,
  title={Multi-task graph-based model for metro flow prediction under dynamic urban conditions},
  author={Zhao, Lu and Zou, Linmu and Wang, Zijia and Song, Taoran and Schonfeld, Paul and Chen, Feng and Li, Rui and Li, Pengcheng},
  journal={Computer-Aided Civil and Infrastructure Engineering},
  volume={40},
  number={21},
  pages={3239--3258},
  year={2025},
  publisher={Wiley Online Library}
}

@article{yang2026cati,
  title={CATI: Cross-attention-based task interaction for multi-granular metro passenger flow forecasting},
  author={Yang, Qiong and Xu, Xianghua and Yu, Juan and Gao, Qifeng and Zhang, Cheng},
  journal={Symmetry},
  volume={18},
  number={5},
  pages={809},
  year={2026},
  publisher={MDPI}
}

\end{document}